\documentclass[journal]{IEEEtran}

\usepackage{setspace}
\usepackage{amsfonts,psfrag}
\usepackage{color}
\usepackage{cite}
\usepackage{graphicx}

\ifCLASSINFOpdf
\else
\fi

\usepackage[cmex10]{amsmath}
\usepackage{array}

\usepackage{url}

\newtheorem{definition}{Definition}
\newtheorem{theorem}{Theorem}
\newtheorem{lemma}{Lemma}
\newtheorem{corollary}{Corollary}
\newtheorem{claim}{Claim}
\newtheorem{example}{Example}

\newcommand{\Qr}{{\mathbb Q}}
\newcommand{\un}{{\em unfixed}}
\newcommand{\fx}{{\em fixed}}

\newcommand{\DD}{{\sc Db-dim}}
\newcommand{\DCv}{{\sc Minor-v}}
\newcommand{\DCe}{{\sc Minor-e}}
\newcommand{\DC}{\DCe}

\newcommand{\DEC}{{\sc Decomp}}
\newcommand{\LOGP}{{\sc Log Partition}}
\newcommand{\MAP}{{\sc Mode}}

\newcommand{\tG}{\widetilde{G}}

\newcommand{\beq}{\begin{eqnarray}}
\newcommand{\eeq}{\end{eqnarray}}
\newcommand{\beqn}{\begin{equation}}
\newcommand{\eeqn}{\end{equation}}
\newcommand{\R}{\mathbb{R}}
\newcommand{\N}{\mathbb{N}}
\newcommand{\Rp}{\mathbb{R}_+}

\newcommand{\beps}{\varepsilon}
\newcommand{\lf}{\left}
\newcommand{\rf}{\right}

\newcommand{\Sg}{\Sigma}

\newcommand{\bdg}{\mathbf{d_G}}

\newcommand{\mc}{\mathcal}
\newcommand{\mb}{\mathbf}

\newcommand{\bdelta}{{\varphi}}
\newcommand{\bDelta}{{\Lambda}}

\newcommand{\cM}{\mc{M}}
\newcommand{\cX}{\mc{X}}

\newcommand{\cR}{\mc{R}}
\newcommand{\cW}{\mc{W}}
\newcommand{\cB}{\mc{B}}
\newcommand{\cG}{\mc{G}}

\newcommand{\cH}{\mc{H}}

\newcommand{\cE}{\mc{E}}
\newcommand{\cT}{\mc{T}}
\newcommand{\cS}{\mc{S}}
\newcommand{\Sgl}{{|\Sigma|}}
\newcommand{\cU}{\mc{U}}

\newcommand{\bB}{\mathbf{B}}
\newcommand{\bX}{\mb{X}}
\newcommand{\bx}{\mb{x}}

\newcommand{\by}{\mb{y}}

\newcommand{\bQ}{\mathbf{Q}}

\newcommand{\hZ}{\hat{Z}}
\newcommand{\hZL}{\hat{Z}_{\text{LB}}}
\newcommand{\hZU}{\hat{Z}_{\text{UB}}}

\newcommand{\bbZ}{\mathbb{Z}}

\newcommand{\pR}{\partial R}

\newcommand{\E}{\mathbb{E}}
\newcommand{\Ex}{{\mathbb E}}
\newcommand{\bbN}{{\mathbb N}}

\newcommand{\Ts}{T_{SAW}}
\newcommand{\Tc}{T_{COMP}}

\newcommand{\vp}{\vspace*{.1in}}
\newcommand{\M}{{\sc Msg-Pass-Mode}}

\begin{document}

\title{Local approximate inference algorithms}
%

\author{Kyomin~Jung and  Devavrat~Shah
\thanks{K. Jung is with the Department of Mathematics and D. Shah is with
the Department of Electrical Engineering and Computer Science,
Massachusetts Institute of Technology, Cambridge, MA 02139 USA.
Email: {\tt \{kmjung,devavrat\}@mit.edu}}
\thanks{This work was  supported in part by a Samsung graduate fellowship,
NSF CAREER grant and DARPA ITMANET grant.}
}

\markboth{Submitted to IEEE Transaction on Information Theory}%
{Shell \MakeLowercase{\textit{et al.}}: Bare Demo of IEEEtran.cls for Journals}

\maketitle

\begin{abstract}

We present a new local approximation algorithm for computing Maximum a Posteriori (MAP) and
log-partition function for arbitrary exponential family distribution
represented by a finite-valued pair-wise Markov random field (MRF),
say $G$. Our algorithm is based on decomposition of $G$ into {\em
appropriately} chosen small components; then computing estimates locally
in each of these components and then producing a {\em good} global
solution. Our algorithm for log-partition function provides provable upper
and lower bounds on the correct value for arbitrary graph $G$. For MAP, our algorithm provides
approximation with quantifiable error for arbitrary $G$. Specifically,
we show that if the underlying graph $G$ either excludes some finite-sized
graph as its minor (e.g. Planar graph) or has low doubling dimension (e.g.
any graph with {\em geometry}), then our algorithm will produce
solution for both questions within {\em arbitrary accuracy}. The
running time of the algorithm is $\Theta(n)$ ($n$ is the number of
nodes in $G$), with constant dependent on  accuracy and either
doubling dimension, or maximum vertex degree and the size of the
graph that  is excluded as a minor (e.g. $3$ for all Planar graphs).

We present a message-passing implementation of our algorithm for
MAP computation using self-avoiding walk of graph. In order to evaluate
the computational cost of this implementation, we derive novel tight bounds
on the size of self-avoiding walk tree for arbitrary graph, which may be
of interest in its own right.

As a consequence of our algorithmic result, we
show that the normalized log-partition function (also known as free-energy) for
a class of {\em regular} MRFs (e.g. Ising model on 2-dimensional grid)
will converge to a limit, that is computable to an arbitrary accuracy, as 
the size of the MRF goes to infinity. This method, like classical sub-additivity
method, is likely to be widely applicable.

\end{abstract}


%
\IEEEpeerreviewmaketitle

\section{Introduction}

Markov Random Field (MRF) \cite{Pea88} based exponential
family of distribution allows for representing distributions in an
intuitive parametric form. Therefore, it has been successful in
modeling many applications (see, \cite{WJ03} for details). The
key operational questions of interest are related
to statistical inference: computing most likely assignment
of (partially) unknown variables given some observations
and computation of probability of an assignment given the
partial observations (equivalently, computing log-partition function).  
In this paper, we study the question of designing efficient  local 
algorithms for solving these inference problems.

\subsection{Previous work}

The question of finding MAP (or ground state) of a given MRF comes up in many important
application areas such as coding theory, 
discrete optimization, 
image denoising. 
Similarly, log-partition function is used in counting combinatorial
objects \cite{SJ89}, loss-probability computation in computer networks, \cite{K91},
etc. Both problems are NP-hard for exact and even (constant)
approximate computation for arbitrary graph $G$. However, the above
stated applications require solving these problems using very 
simple algorithms. A popular successful approach for designing efficient
heuristics has been as follows. First,  identify a wide class of graphs that 
have simple algorithms for computing MAP and log-partition function. 
Then, for any given graph,  approximately compute solution either
by using that simple algorithm as a heuristic or in a more sophisticated
case,  by possibly solving multiple sub-problems induced by sub-graphs with 
good graph structures and then combining the results from these sub-problems 
to obtain a global solution.

Such an approach has resulted in many interesting recent results starting
the Belief Propagation (BP) algorithm designed for Tree graph \cite{Pea88}.
Since there is a vast literature on this topic, we will recall only few results.
In our opinion, two important algorithms proposed along these lines of thought 
are the generalized belief propagation (BP)  \cite{YFW00}
and the tree-reweighted algorithm (TRW) \cite{WJWa,WJWb, WJWc}.
Key properties of interest for these iterative procedures
are the correctness of their fixed points and convergence. Many
results characterizing properties of the fixed points are known 
starting from \cite{YFW00}. 
Various sufficient conditions for their convergence are known 
starting \cite{TJ02}. However, simultaneous convergence and 
correctness of such algorithms are established for only specific 
problems, e.g. \cite{BSS05, MV06, KW06}.

Finally, we discuss two relevant results. The first result is about
properties of TRW. The TRW algorithm provides provable upper bound
on log-partition function for arbitrary graph \cite{WJWc}.
However, to the best of authors' knowledge the error is not quantified.
The TRW for MAP estimation has a strong connection to specific Linear
Programming (LP) relaxation of the problem \cite{WJWb}.
This was made precise in a sequence of work by Kolmogorov \cite{Kol05},
Kolmogorov and Wainwright \cite{KW06} for binary MRF. It is worth noting
that LP relaxation can be poor even for simple problems.

The second is an approximation algorithm proposed by Globerson and
Jaakkola \cite{GJ06} to compute log-partition function using
Planar graph decomposition (PDC). PDC uses techniques of \cite{WJWc}
in conjunction with known result about exact computation of
partition function for binary MRF when $G$ is Planar and the
exponential family has a specific form (binary pairwise and multiplicative
potentials). Their algorithm provides provable upper bound for 
arbitrary graph. However, they do not quantify the error incurred. 
Further, their algorithm is limited to binary MRF.

\subsection{Contributions}

We propose a novel local algorithm for approximate computation of MAP and log-partition
function. For any $\varepsilon > 0$, our algorithm can produce an
$\beps$-approximate solution for MAP and log-partition function for
{\em arbitrary} MRF $G$ as long as $G$ has either of these two properties:
(a) $G$ has low doubling dimension (see Theorems \ref{thm:main0} and \ref{thm:main0x}),
or (b) $G$ excludes a finite-sized graph as a minor
(see Theorems \ref{thm:main1} and \ref{thm:main1x}).
For example, Planar graph excludes $K_{3,3}, K_5$ as a minor and thus
our algorithm provides approximation algorithms for Planar graphs.

The running time of the algorithm is $\Theta(n)$, with constant dependent
on $\beps$ and (a) doubling dimension for doubling dimension graph, or
(b) maximum vertex degree and size of the graph that is excluded as minor
for minor-excluded graphs. For example, for $2$-dimensional grid
graph, which has doubling dimension $O(1)$, the algorithm takes
$C(\beps) n$ time, where $\log \log C(\beps) = O(1/\beps)$.  On the other
hand, for a planar graph with maximum vertex degree a constant, i.e. $O(1)$,
the algorithm takes $C'(\beps) n$ time, with $\log \log C'(\beps) = O(1/\beps)$.

In general, our algorithm works for any $G$ and we can quantify bound
on the error incurred by our algorithm. It is worth noting that our algorithm
provides a provable lower bound on log-partition function as well unlike
many of the previous results.

Our algorithm is primarily based on the following idea: First,
decompose $G$ into small-size connected components say
$G_1,\dots, G_k$ by removing few edges of $G$. Second,
compute estimates (either MAP or log-partition) in each of the $G_i$
separately. Third, combine these estimates to produce a global
estimate while {\em taking care} of the {\em error} induced by the removed
edges. We show that the error in the estimate depends only
on the edges removed. This error bound characterization is
applicable for arbitrary graph.

For obtaining sharp error bounds,
we need good graph decomposition schemes. Specifically, we
use a new, simple and very intuitive randomized decomposition
scheme for graphs with low doubling dimensions. For minor-excluded
graphs, we use a simple scheme based on work by Klein, Plotkin
and Rao \cite{KPR94} and Rao \cite{R99} that they had introduced
to study the gap between max-flow and min-cut for multicommodity
flows.  In general, as long as $G$ allows for such good edge-set for
decomposing $G$ into small components, our algorithm will provide a
good estimate.

To compute estimates in individual components, we use
dynamic programming. Since each component is small, it is not
computationally burdensome.  However, one may obtain further simpler
heuristics by replacing dynamic programming by other method such as
BP or TRW for computation in the components.

In order to implement dynamic programing using message-passing
approach, we use construction based on self-avoiding walk tree.
Self-avoiding walk trees have been of interest in statistical physics for
various reasons (see book by Madras and Slade \cite{MS93}).
Recently, Weitz \cite{DW06} obtained a surprising result that connected
computation of marginal probability of a node in any binary MRF to
that of marginal probability of a root node in an appropriate self-avoiding
walk tree. We use a direct adaption of this result for computing MAP
estimate to design message passing scheme for MAP computation.
In order to evaluate computation cost, we needed tight bound on the
size on self-avoiding walk tree of arbitrary graph $G$. We obtain a
novel characterization of size of self-avoiding walk tree within a
factor $8$ for arbitrary graph $G$. This result should be of interest
in its own right.

Finally, as a (somewhat unexpected) consequence of these algorithmic results,
we obtain a method to establish existence of asymptotic limits of
free energy for a class of MRF. Specifically, we show that if
the MRF is $d$-dimensional grid and all node, edge potential functions
are identical then the free-energy (i.e. normalized log-partition function)
converges to a limit as the size of the grid grows to infinity. In general,
such approach is likely to extend for any {\em regular enough}
MRF for proving existence of such limit: for example, the result
will immediately extend to the case when the requirement of 
node, edge potential being exactly the same is replaced by the
requirement of they being chosen from a common distribution in 
an i.i.d. fashion.

\subsection{Outline}

The paper is organized as follows. Section \ref{sec:prelim} presents
necessary background on graphs, Markov random fields, exponential
family of distribution, MAP estimation and log-partition function
computation.

Section \ref{sec:decomp} presents graph decomposition schemes.
These decomposition schemes are used later by approximation
algorithms. We present simple, intuitive and $O(n)$ running time
decomposition schemes for graphs with low doubling dimension
and graphs that exclude finite size graph as a minor. Both of these
schemes are randomized. The first scheme is our original contribution.
The second scheme was proposed by Klein, Plotkin and Rao \cite{KPR94}
and Rao \cite{R99}.

Section \ref{sec:logp} presents the approximation algorithm for
computing log-partition function. We describe how it provides upper
and lower bound on log-partition function for arbitrary graph.
Then we specialize the result for two graphs of interest: low
doubling dimension and minor excluded graphs.

Section \ref{sec:map} presents the approximation algorithm for
MAP estimation. We describe how it provides approximate estimate
for arbitrary graph with quantifiable approximation error.  Then we
specialize the result for two graphs of interest: low doubling
dimension and minor excluded graphs.

Section \ref{sec:mp} describes message passing implementation
of the MAP estimation algorithm for binary pair-wise MRF for
arbitrary $G$. This can be used by our approximation algorithm
to obtain message passing implementation. This algorithm builds
upon work by Weitz \cite{DW06}.  We describe a novel tight bound on
the size of self-avoiding walk tree for any $G$. This helps in evaluating
the computation time. The message passing implementation
has similar computation complexity as the centralized algorithm.

Section \ref{sec:exp} presents an experimental evaluation of our
algorithm for popular synthetic model on  a grid graph. We compare
our algorithm with TRW and PDC algorithms and show that our
algorithm is very competent. An important feature of our algorithm
is scalability.

Section \ref{sec:imp} presents the implication of our algorithmic
result in establishing asymptotic limit of free energy for  regular
MRFs, such as an  Ising model on $d$-dimensional grid.

\subsection{How to read this paper: a suggestion}

A reader, interested in obtaining a quick understanding of the results, should
skip everything in Section \ref{sec:decomp}  other than the definition
of $(\beps, \Delta)$ decomposition, and skip the
Section \ref{sec:mp} completely. Reading these two sections
at the very end may be helpful to parse the results with ease for
all the readers.

\section{Preliminaries}\label{sec:prelim}

This section provides the background necessary for subsequent
sections. We begin with an overview of some graph theoretic
basics. We then describe formalism of Markov random field and
exponential family of distribution. We formulate the problem
of log-partition function computation and MAP estimation for
Markov random field. We conclude by stating precise definitions
of approximate MAP estimation and approximate log-partition
function computation.

\subsection{Graphs}

An undirected graph $G=(V,E)$ consists of a set
of vertices $V=\{1,\dots, n\}$ that are connected
by set of edges $E \subset V \times V$. We consider
only simple graphs, that is multiple edges between
a pair of nodes or self-loops are not allowed. Let
$\Gamma(v) = \{u \in V : (u,v) \in E\}$ denote the
set of all neighboring nodes of $v \in V$. The size
of the set $\Gamma(v)$ is the {\em degree} of node $v$,
denoted as $d_v$. Let $d^* = \max_{v \in V} d_v$ be the
maximum vertex degree in $G$. A {\em clique} of the graph $G$ is
a fully-connected subset $C$ of the vertex set
(i.e. $(u,v) \in E$ for all $u, v \in C$). Nodes
$u$ and $v$ are called  connected if there
exists a path in $G$ starting  from $u$ and
ending at $v$ or vice versa since $G$
is undirected. Each graph $G$ naturally decomposes
into disjoint sets of vertices $V_1,\dots, V_k$ where
for $1\leq i\leq k$, any two nodes say $u, v \in V_i$ are
connected. The sets $V_1,\dots, V_k$ are called the
connected components of $G$.

We introduce a popular notion of {\em dimension} for
graph $G$ (see recent works \cite{KR02} \cite{GKL03}
\cite{HM04} for relevant details). Define $\bdg: V \times V \to \Rp$ as
$$\bdg(i,j) = \mbox{\sf length of the shortest path between $i$ and $j$}.$$
If $i = j$ then $\bdg(i,j) =0$ and if $i, j$ are not connected,
then define $\bdg(i,j) = \infty$. It is easy to
check that thus defined $\bdg$ is a metric on vertex set $V$.
Define ball of radius $r \in \Rp$ around $v \in V$ as
$\bB(v, r) = \{ u \in V : \bdg(u,v) < r\}$.
Define
\begin{eqnarray*}
\rho(v, r) & = & \inf \{ K \in \N:  \exists ~u_1,\dots, u_K \in V, \\
& ~ & \qquad \qquad \bB(v,r) \subset \cup_{i=1}^K \bB(u_i, r/2) \}.
\end{eqnarray*}
Then, $\rho(G) = \sup_{v \in V, r \in \Rp} \rho(v,r)$ is called
the {\em doubling dimension} of graph $G$. Intuitively, this
definition captures the notion of dimension $d$ in the Euclidian
space $\R^d$. It follows from definition that for any graph $G$,
$\rho(G) = O(\log_2 n)$. We note the following property whose
proof is presented in Appendix \ref{ap:01}.
\begin{lemma}\label{lem:poly} For any $v \in V$ and $r \in \N$,
$ | \bB(v, 2^r) | \leq 2^{r\rho(\cM)}.$
\end{lemma}

Next, we introduce a class of graphs known as {\em minor-excluded}
graphs (see a series of publications by Roberston and Seymour under
"the graph minor theory" project \cite{RS}). A graph $H$ is called minor of $G$ if we can transform
$G$ into $H$ through an arbitrary sequence of the following two operations: (a)
removal of an edge; (b) merge two connected vertices $u,v$: that is,
remove edge $(u,v)$ as well as vertices $u$ and $v$; add a new
vertex and make all edges incident on this new vertex that were
incident on $u$ or $v$. Now, if $H$ is not a minor of $G$ then we
say that $G$ excludes $H$ as a minor.

The explanation of the following statement may help understand the
definition better: {\em any graph $H$ with $r$ nodes is a minor of
$K_{r}$}, where $K_{r}$ is a complete graph of $r$ nodes. This is
true because one may obtain $H$ by removing edges from $K_r$ that
are absent in $H$. More generally, if $G$ is a subgraph of $G'$ and
$G$ has $H$ as a minor, then $G'$ has $H$ as its minor. Let
$K_{r,r}$ denote a complete bipartite graph with $r$ nodes in each
partition. Then $K_r$ is a minor of $K_{r,r}$. An important
implication of this is as follows: to prove property {\cal P} for
graph $G$ that excludes $H$, of size $r$, as a minor, it is
sufficient to prove that any graph that excludes $K_{r,r}$ as a
minor has property {\cal P}. This fact was cleverly
used by Klein et. al. \cite{KPR94}.

\subsection{Markov random field}

A Markov Random Field (MRF) is defined on
the basis of an undirected graph $G=(V,E)$ in the
following manner. Let $V = \{1,\dots, n\}$ and
$E \subset V \times V$. For each $v \in V$, let $X_v$
be random variable taking values in some finite
valued space $\Sigma_v$. Without loss of generality,
lets assume that $\Sigma_v = \Sigma$ for all $v \in V$.
Let $\bX = (X_1,\dots, X_n)$ be the collection of
these random variables taking values in $\Sigma^n$.
For any subset $A \subset V$, we let
$\bX_A$ denote $\{X_v | v \in A\}$. We call a subset
$S \subset V$ a {\em cut} of $G$ if by its removal from
$G$ the graph decomposes into two or more disconnected
components. That is, $V \backslash S = A \cup B$
with $A \cap B = \emptyset$ and there for any $a \in A, b \in B$,
$(a,b) \notin E$. The $\bX$ is called a Markov random field, if for any cut
$S \subset V$, $\bX_A$ and $\bX_B$ are conditionally
independent given $\bX_S$, where $V \backslash S = A \cup B$.

By the Hammersley-Clifford theorem, any Markov random
field that is strictly positive (i.e. $\Pr(\bX=\bx) > 0$ for
all $\bx \in \Sigma^n$) can be defined in terms of a
decomposition of the distribution over cliques of the graph.
Specifically, we will restrict our attention to pair-wise
Markov random fields (to be defined precisely soon) only
in this paper. This does not incur loss of generality
for the following reason. A distributional
representation that decomposes in terms of distribution
over cliques can be represented through a factor graph
over discrete variables. Any factor graph over discrete
variables can be tranformed into a  pair-wise
Markov random field (see, \cite{WJWb} for example) by introducing
auxiliary variables. As reader shall notice, the
techniques of this paper can be extended to Markov
random fields with higher-order interaction that contains
hyper-edges.

Now, we present the precise definition of pair-wise Markov random
field. We will consider distributions in exponential form.
For each vertex $v \in V$ and edge $(u,v) \in E$, the corresponding
potential functions are $\phi_v : \Sigma \to \Rp$ and
$\psi_{uv}: \Sigma^2 \to \Rp$. Then, the distribution of $\bX$
is given as follows: for $\bx \in \Sigma^n$,
\beq \Pr[\bX=\bx] &
\propto & \exp\lf(\sum_{v \in V} \phi_v(x_v) + \sum_{(u,v)\in
E}\psi_{uv}(x_u,x_v)\rf). ~~\label{markovicity}
\eeq
We note that the assumption of $\phi_v, \psi_{uv}$ being non-negative
does not incur loss of generality for the following reasons: (a) the
distribution remains the same if we consider potential
functions $\phi_v + C, \psi_{uv} + C$, for all $v \in V, (u,v) \in E$
with constant $C$; and (b)
by selecting large enough constant, the modified functions will become
non-negative as they are defined over finite discrete domain.

\subsection{Log-partition function}

The normalization constant in definition (\ref{markovicity}) of
distribution is called the {\em partition function}, $Z$. Specifically,
$$ Z = \sum_{\bx \in \Sigma^n} \exp\left(\sum_{v \in V} \phi_v(x_v) + \sum_{(u,v)\in
E}\psi_{uv}(x_u,x_v)\rf).$$
Clearly, the knowledge of $Z$ is necessary in order to evaluate
probability distribution or to compute marginal probabilities,
i.e. $\Pr(X_v = x_v)$ for $v \in V$. In applications in
computer science, $Z$ corresponds to the number of
combinatorial objects, in statistical physics normalized
logarithm of $Z$ provides free-energy and in reversible
stochastic networks $Z$ provides loss probability for
evaluating quality of service.

In this paper, we will be interested in obtaining
estimate of $\log Z$. Specifically, we
will call $\hZ$ as an $\beps$-approximation of $Z$ if
$$ (1-\beps) \log Z \leq \log \hZ \leq (1+\beps) \log Z.$$

\subsection{MAP assignment}

The maximum a posteriori (MAP) assignment
$\bx^*$ is one with maximal probability, i.e.
$$\bx^* \in \arg\max_{\bx \in \Sigma^n} \Pr[\bX=\bx].$$
Computing MAP assignment is of interest in
wide variety of applications. In combinatorial
optimization problem, $\bx^*$ corresponds to
an optimizing solution, in the context of image
processing it can be used as the basis for
image segmentation techniques and in error-correcting
codes it corresponds to decoding the received
noisy code-word.

In our setup, MAP assignment $\bx^*$ corresponds to
$$ \bx^* \in \arg\max_{\bx \in \Sigma^n} \lf(\sum_{v \in V} \phi_v(x_v) +
 \sum_{(u,v) \in E} \psi_{uv}(x_u, x_v)\rf).$$
Define, $\cH(\bx) = (\sum_{v \in V} \phi_v(x_v) +
 \sum_{(u,v) \in E} \psi_{uv}(x_u, x_v))$.
We will be interested in obtaining an $\beps$ estimate, say $\widehat{\bx}$, of
$\bx^*$ such that
$$ (1-\beps) \cH(\bx^*) \leq \cH(\widehat{\bx}) \leq \cH(\bx^*).$$

\section{Graph decomposition}\label{sec:decomp}

In this section, we introduce notion of
{\em graph decomposition}. We describe
very simple algorithms for obtaining
decomposition for graphs with low doubling
dimension and minor-excluded graphs. In
the later sections, we will show that such
decomposable graphs are good structures in
the sense that they allow for local algorithms
for approximately computing log-partition
function and MAP.

\subsection{$(\beps, \Delta)$ decomposition}

Given $\beps, \Delta > 0$, we define notion
of $(\beps, \Delta)$ decomposition for a graph
$G = (V,E)$. This notion can be stated in terms
of vertex-based decomposition or edge-based
decomposition.

We call a random subset of vertices $\cB \subset V$
as $(\beps, \Delta)$ vertex-decomposition of $G$ if
the following holds:
(a) For any $v \in V$, $\Pr(v \in \cB) \leq \beps$.
(b) Let $S_1,\dots,S_K$ be connected components of
graph $G'=(V', E')$ where $V' = V \backslash \cB$
and $E' = \{(u,v) \in E: u, v \in V'\}$. Then,
$\max_{1\leq k\leq K} |S_k| \leq \Delta$ with probability
$1$.

Similarly, a random subset of edges $\cB \subset E$
is called an $(\beps, \Delta)$ edge-decomposition of $G$ if
the following holds:
(a) For any $e \in E$, $\Pr(e \in \cB) \leq \beps$.
(b) Let $S_1,\dots,S_K$ be connected components of
graph $G'=(V', E')$ where $V' = V$ and $E' = E \backslash \cB$.
Then,  $\max_{1\leq k\leq K} |S_k| \leq \Delta$ with probability
$1$.

\subsection{Low doubling-dimension graphs}

This section presents $(\beps, \Delta)$ decomposition algorithm
for graphs with low doubling dimension for various choice of
$\beps$ and $\Delta$. Such a decomposition algorithm can be
obtained through a probabilistically padded decomposition
for such graphs \cite{GKL03}. However, we present our (different)
algorithm due to its simplicity. Its worth noting that this
simplicity of the algorithm requires proof technique different (and
more complicated) than that known in the literature.

We will describe algorithm for node-based $(\beps,\Delta)$ decomposition.
This will immediately imply algorithm for edge-based decomposition for
the following reason: given $G=(V,E)$ with doubling dimension $\rho(G)$,
consider graph of its edges $\cG = (E,\cE)$ where $(e, e') \in \cE$
if $e, e'$  shared a vertex in $G$. It is easy to check that
$\rho(\cG) \leq 2\rho(G) + 1$. 
Therefore, running algorithm for node-based decomposition
on $\cG$ will provide an edge-based decomposition.

The node-based decomposition algorithm for $G$ will be described
for the metric space on $V$ with respect to the shortest
path metric $\bdg$ introduced earlier. Clearly, it is not
possible to have $(\beps, \Delta)$ decomposition for any $\beps$
and $\Delta$ values. As will become clear later, it is important
to have such decomposition for $\beps$ and $\Delta$ being not
too large (specifically, we would like $\Delta = O(\log n)$).
Therefore, we describe algorithm for any $\beps > 0$ and an operational
parameter $K$. We will show that the algorithm will produce
$(\beps, \Delta)$ node-decomposition where $\Delta$ will depend
on $\beps, K$ and $\rho$.

Given $\beps$ and $K$, define random variable $\bQ$ over
$\{1,\dots, K\}$ as
$$ \Pr[\bQ = i] = \begin{cases} \beps (1-\beps)^{i-1} & \mbox{if $1\leq i < K$} \\
(1-\beps)^{K-1}  & \mbox{if $i = K$}
\end{cases}.$$
Define, $P_K = (1-\beps)^{K-1}$. The algorithm \DD($\beps, K$) described
next essentially does the following: initially, all vertices are
colored {\em white}. Iteratively, choose a white vertex {\em arbitrarily}. Let
$u_t$ be vertex chosen in iteration $t$. Draw an independent random number
as per distribution of $\bQ$, say $\bQ_t$. Select all {\em white} vertices
that are at distance $\bQ_t$ from $u_t$ in $\cB$ and color them {\em blue}; color
all {\em white} vertices at distance $< \bQ_t$ from
$u_t$ (including itself) as {\em red}. Repeat this process till no
more {\em white} vertices are left. Output $\cB$ (i.e. {\em blue} nodes)
as the decomposition. Now, precise description of the algorithm.

\vspace{.05in}
\noindent{\bf \DD ($\beps, K$)}
\vspace{.05in}
\hrule
\vspace{.1in}
\begin{itemize}
\item[(1)] Initially, set iteration number $t=0$, $\cW_0=V$, $\cB_0=\emptyset$ and $\cR_0=\emptyset$.
\item[(2)] Repeat the following till $\cW_t \neq \emptyset$:
\begin{itemize}
\item[(a)] Choose an element $u_{t} \in \cW_t$ uniformly at random.
\item[(b)] Draw a random number $Q_t$ independently according to the distribution of $\bQ$.
\item[(c)] Update
\begin{itemize}
\item[(i)] $ \cB_{t+1} \leftarrow \cB_t \cup \{w | \bdg(u_t,w)= Q_t ~\mbox{ and }~ w \in \cW_t\}$,
\item[(ii)] $\cR_{t+1} \leftarrow \cR_t \cup \{w | \bdg(u_t,w)< Q_t ~\mbox{ and }~ w\in \cW_t\}$,
\item[(iii)] $\cW_{t+1} \leftarrow \cW_t \cap \left(\cB_{t+1} \cup \cR_{t+1}\right)^c$.
\end{itemize}
\item[(d)] Increment $t \leftarrow t+1$.

\end{itemize}
\item[(3)] Output $\cB_t$.
\end{itemize}
 \vspace{.1in} \hrule \vspace{.05in}

We state property of the algorithm \DD($\beps, K$) as
follows.
\begin{lemma}\label{lem:decomposition}
Given $G$ with doubling dimension $\rho = \rho(G)$ and $
\beps\in (0,1)$, let $ K(\beps, \rho) = \frac{12\rho}{\beps} \log
\lf(\frac{24\rho}{\beps}\rf).$ Then \DD($\beps, K(\beps, \rho)$)
produces random output $\cB \subset V$ that is $(2\beps, \Delta(\beps, \rho))$
node-decomposition of $G$ with $\Delta(\beps, \rho) \leq K(\beps, \rho)^{2\rho}$.
The algorithm takes $O(C(\beps, \rho) n)$ amount of time to produce $\cB$,
where $C(\beps, \rho) = K(\beps, \rho)^{2\rho}$.
\end{lemma}

Before presenting the proof of Lemma \ref{lem:decomposition}, we
state the following important corollary for designing efficient
algorithm.
\begin{corollary}\label{cor:decomposition}
Let $\beps \leq 1, \rho$ be such that $\rho \log (\rho/\beps) = o(\log \log n)$.
Then \DD($\beps/2, K(\beps/2, \rho)$) produces $(\beps, \log^{1/L} n)$
node-decomposition for any finite (not scaling with $n$) $L$ .
\end{corollary}
\begin{proof}
Since $\rho \log (\rho/\beps) = o(\log \log n)$, we have that
$$ 2\rho \lf(\log \frac{24\rho}{\beps} + \log \log \frac{48\rho}{\beps}\rf) = o(\log \log n).$$
Therefore, by definition of $K(\beps, \rho)$ we have that
\beq
K(\beps/2,\rho)^{2\rho} & = & \exp\lf(2\rho \lf[\log \frac{24\rho}{\beps} + \log \log \frac{48\rho}{\beps}\rf]\rf) \nonumber \\
& = & \exp\lf(o(\log \log n)\rf) \nonumber \\
& \leq & \log^{1/L} n,
\eeq
for any finite $L$. The last inequality follows from the definition of
notation $o(\cdot)$. Now, Lemma \ref{lem:decomposition} implies the desired claim.
\end{proof}

\begin{proof}{\em (Lemma \ref{lem:decomposition})} To
prove claim of Lemma, we need to show two properties
of the output set $\cB$ for given $\beps$ and
$K = K(\beps, \rho)$: (a) for any $v \in V$,
$\Pr(v \in \cB) \leq 2\beps$; (b) the graph $G$, upon
removal of $\cB$, decomposes into connected component
each of size at most $K^{2\rho}$. 

Before, we prove (a) and (b), lets bound the running time
of the algorithm. Note that the algorithm runs for at most $n$ 
iterations. In each iteration, the algorithm needs to check
nodes that are within distance $K(\beps, \rho)$ of the 
randomly chosen node. Therefore, total number of operations
performed is at most $O(K(\beps,\rho)^{2\rho})$. Thus, the
total running time is $O(n K(\beps,\rho)^{2\rho})$. Now 
we first justify (a) and then (b).

\vp
\noindent{\em Proof of (a).} To prove (a), we use the following Claim.
\begin{claim}\label{claim:two}
Consider metric space $\cM = (V, \bdg)$ with $|V|=n$.  Let $\cB
\subset V$ be the random set that is output of decomposition
algorithm with parameter $(\beps, K)$ applied to $\cM$.
Then, for any $v \in V$
$$ \Pr[v \in \cB] \leq \beps +  P_K | \bB(v, K)|, $$
where $\bB(v,K)$ is the ball of radius $K$ in $\cM$ with respect to
the $\bdg$.
\end{claim}
\begin{proof}{\em (Claim \ref{claim:two})}
The proof is by induction on the number of points $n$ over which
metric space is defined.   When $n=1$, the algorithm chooses only
point as $u_0$ in the initial iteration and hence it can not be part
of the output set $\cB$. That is, for this only point, say $v$,
$$\Pr[v\in \cB] = 0 \leq \beps + P_K |\bB(v,K)|.$$
Thus, we have verified the base case for induction ($n=1$).

As induction hypothesis, suppose that the Claim \ref{claim:two} is
true for any metric space on $n$ points with $n< N$ for some $N \geq
2$. As the induction step, we wish to establish that for a metric
space $\cM=(V, \bdg)$ with $|V| = N$, the Claim \ref{claim:two} is
true. For this, consider any point $v \in V$. Now consider the first
iteration of the algorithm applied to $\cM$. The algorithm picks
$u_0 \in V$ uniformly at random in the first iteration. Given $v$,
depending on the choice of $u_0$ we consider four different cases
(or events).

\vp

{\em Case 1.}   This case corresponds to event $E_1$ where the
chosen random $u_0$ is equal to point $v$ of our interest. By
definition of algorithm, under the event $E_1$, $v$ will never be
part of output set $\cB$. That is,
$$ \Pr[v\in \cB|E_1] = 0 ~ \le~ \beps + P_K |\bB(v,K)|.$$

\vp

{\em Case 2.} Now, suppose $u_0$ be such that $v \neq u_0$ and
$\bdg(u_0, v) < K$. Call this event $E_2$. Further, depending on
choice of random number $Q_0$, define events:
$$ E_{21}= \{\bdg(u_0,v) < Q_0\}, ~ E_{22}=\{\bdg(u_0,v) = Q_0\}, ~\mbox{and}~$$ $$ ~E_{23}=\{\bdg(u_0,v) > Q_0\}.$$
By definition of algorithm, when $E_{21}$ happens, $v$ is selected
as part of $\cR_1$ and hence can never be part of output $\cB$. When
$E_{22}$ happens $v$ is selected as part of $\cB_1$ and hence it is
definitely part of output set $\cB$. When $E_{23}$ happens, $v$ is
neither selected in set $\cR_1$ nor selected in set $\cB_1$. It is
left as an element of the set $\cW_1$.  This new set $\cW_1$ has
points $< N$. The original metric $\bdg$ is still a metric on
points\footnote{Note the following subtle but crucial point. We are
not changing the metric $\bdg$ after we remove points from original
set of points as part of the algorithm.} of $\cW_1$.  By definition,
the algorithm only cares about $(\cW_1, \bdg)$ in future and is not
affected by its decisions in past. Therefore, we can invoke
induction hypothesis which implies that if event $E_{23}$ happens
then the probability of $v \in \cB$ is bounded above by $ \beps +
P_K |\bB(v, K)|$. Finally, let us relate the $\Pr[E_{21}|E_2]$ with
$\Pr[E_{22}|E_2]$. Suppose $\bdg(u_0, v) = \ell < K$. By definition
of probability distribution of $\bQ$, we have \beq
 \Pr[E_{22}|E_2]   & = &  \beps (1-\beps)^{\ell-1}.
 \eeq
 \beq
 \Pr[E_{21}|E_2]   & = &  (1-\beps)^{K-1} + \sum_{j=\ell+1}^{K-1} \beps (1-\beps)^{j-1} \nonumber \\
  & = & (1-\beps)^{\ell}.
  \eeq
That is,
 $$ \Pr[E_{22}|E_2] = \frac{\beps}{1-\beps} \Pr[E_{21}|E_2].$$
Let $q \stackrel{\triangle}{=} \Pr[E_{21}|E_2]$.
Then,
\beq
\Pr[v \in \cB | E_2] & = & \Pr[v \in \cB|E_{21} \cap E_2] \Pr[E_{21}|E_2] \nonumber \\
& ~& ~~ + \Pr[v\in \cB|E_{22} \cap E_2]\Pr[E_{22}|E_2] \nonumber \\
& ~ & ~~~+ \Pr[v\in \cB|E_{23} \cap E_2 ]\Pr[E_{23}|E_2] \nonumber \\
& = & \frac{\beps q}{1-\beps}  + (\beps + P_K |\bB(v,K)|) \lf(1-\frac{q}{1-\beps}\rf) \nonumber \\
& = & \beps + P_K |\bB(v,K)| - \frac{q P_K|\bB(v,K)|}{1-\beps} \nonumber \\
& \leq &   \beps + P_K |\bB(v,K)|. \eeq

\vp

{\em Case 3.} Now, suppose $u_0 \neq v$ is such that $\bdg(u_0, v) =
K$. Call this event $E_3$.  Further, let event $E_{31}=\{Q_0 = K\}$.
Due to independence of selection of $Q_0$,  $\Pr[E_{31} | E_3] =
P_K$. Under event $E_{31} \cap E_3$, $v \in \cB$ with probability
$1$. Therefore,
\beq
\Pr[v \in \cB | E_3] & = & \Pr[v \in \cB|E_{31} \cap E_3] \Pr[E_{31}|E_3] \nonumber \\
& ~ & + \Pr[v \in \cB|E^c_{31} \cap E_3] \Pr[E^c_{31}|E_3] \nonumber \\
& = & P_K + \Pr[v \in \cB | E_{31}^c \cap E_3] (1-P_K).~~ \label{xx1}
\eeq
Under event, $E^c_{31} \cap E_3$, we have $v \in
\cW_1$ and the remaining metric space $(\cW_1, \bdg)$. This metric
space has $< N$ points. Further, the ball of radius $K$ around $v$
with respect to this new metric space has at most $|\bB(v, K)| - 1$
points (this ball is with respect to the original metric space $\cM$
of $N$ points). We can invoke induction hypothesis for this new
metric space (because of similar justification as in the previous
case) to obtain \beq \Pr[v \in \cB | E^c_{31} \cap E_3] & \leq &
\beps + P_K (|\bB(v,K)| - 1). \label{xx2} \eeq From (\ref{xx1}) and
(\ref{xx2}), we have \beq
\Pr[v \in \cB | E_3] & \leq & P_K + (1-P_K) (\beps + P_K (|\bB(v,K)| - 1)) \nonumber \\
& = &  \beps (1-P_K) + P_K |\bB(v,K)| \nonumber \\
& ~ & ~~ + P_K^2 (1- |\bB(v,K)|) \nonumber \\
& \leq & \beps + P_K |\bB(v,K)|. \eeq

\vp

{\em Case 4.} Finally, let $E_4$ be the event that $\bdg(u_0, v) >
K$. Then, at the end of the first iteration of the algorithm, we
again have the remaining metric space $(\cW_1, \bdg)$ such that
$|\cW_1| < N$. Hence, as before by induction hypothesis we will have
$$\Pr[v\in \cB|E_4] \le \beps + P_K |\bB (v,K)|.$$

Now, the four cases are exhaustive and disjoint. That is,
$\cup_{i=1}^4 E_i$ is the universe. Based on the above discussion,
we obtain the following. \beq
\Pr[v \in \cB] & = & \sum_{i=1}^4 \Pr[ v \in \cB | E_i] \Pr[E_i] \nonumber \\
 & \leq & \lf(\max_{i=1}^4 \Pr[ v \in \cB | E_i]\rf) \lf(\sum_{i=1}^4 \Pr[E_i] \rf) \nonumber \\
 & \leq & \beps + P_K |\bB(v,K)|.
 \eeq
This completes the proof of Claim \ref{claim:two}.
\end{proof}
Now, we will use Claim \ref{claim:two} to complete the proof of (a).
 Lemma \ref{lem:poly} for metric space with doubling dimension $\rho$
and integer distances imply that,
$$ \lf|\bB(v, K)\rf| \leq \lf|\bB\lf(v, 2^{\lceil \log_2 K\rceil}\rf)\rf| \leq 2^{\rho (\log_2 K + 1)} = (2K)^\rho.$$
Therefore,  it is sufficient to show that
$$ P_K (2K)^\rho \leq \beps.$$
Recall that $ K(\beps, \rho) = \frac{12\rho}{\beps} \log
\lf(\frac{24\rho}{\beps}\rf),$ and $P_K=(1-\beps)^{K-1}$. Hence,
\beq
K  & = & \frac{12\rho}{\beps}\log \lf(\frac{24\rho}{\beps}\rf) \nonumber \\
& \geq & \frac{6\rho}{\beps}\lf( \log \lf(\frac{24\rho}{\beps}\rf) + \log
\log \lf(\frac{24\rho}{\beps}\rf)\rf) \nonumber \\
& = & \frac{6\rho}{\beps}\log 2K.
\eeq
 Now since $K\ge 3$, we obtain that $K-1 \ge
\frac{4\rho}{\beps}\log 2K$. Then, from $K\ge \frac{1}{\beps}$ and
$\rho\ge 1$, $$K-1 \ge \frac{2\rho}{\beps}\log 2K +\frac{2}{\beps}
\log \frac{1}{\beps}.$$ Note that $\log (1-\beps)^{-1}\ge \log
(1+\beps) \ge \frac{\beps}{2},~ ~ \mbox{for} ~ \beps\in (0,1).$
Hence,
$$(K-1)\log (1-\beps)^{-1} \ge \rho \log 2K + \log \frac{1}{\beps},$$
which implies
$$(1-\beps)^{K-1}(2K)^\rho \le \beps.$$
This completes the proof of (a) of Lemma  \ref{lem:decomposition}.

\vp {\em Proof of (b).}
First we give some notations.  Define $R_t = \cR_t - \cR_{t-1}$,
$B_t = \cB_t - \cB_{t-1}$ and
$$\pR_t = \{v \in V:  v \notin R_t \mbox{~and~} \exists~ v' \in R_t \mbox{~s.t.~} \bdg(v, v') = 1\}.$$
The followings are straightforward observations implied by the
algorithm: for any $t \geq 0$, (i) $R_t \cap \cR_{t-1} = \emptyset$,
(ii) $B_t \cap \cB_{t-1} = \emptyset$, (iii) $R_t \subset
\bB(u_{t-1}, Q_{t-1})$, and (iv) $B_t \subset \bB(u_{t-1},
Q_{t-1}+1) - \bB(u_{t-1}, Q_{t-1})$. Now, we state a crucial claim
for proving (b).
\begin{claim}\label{claim:one}
For all $t \geq 0$, $\pR_t \subset \cB_t$.
\end{claim}
\begin{proof}{\em (Claim \ref{claim:one})}
We prove it by induction. Initially, $\pR_0 = \cB_0 = \emptyset$ and
hence the claim is trivial.  At the end of the first iteration, by
definition of the algorithm
$$ R_1 = \cR_1 = \bB(u_0, Q_0), ~\mbox{and}~ $$
$$ B_1 = \cB_1 = \bB(u_0, Q_0+1) - \bB(u_0, Q_0).$$
Therefore, by definition $\pR_1 = \cB_1$.  Thus, the base case of
induction is verified. Now, as  the hypothesis for induction suppose
that $\pR_t \subset \cB_t$ for all $t \leq \ell$, for some $\ell
\geq 1$. As induction step, we will establish that $\pR_{\ell+1}
\subset \cB_{\ell+1}$.

Suppose to the contrary, that $\pR_{\ell +1}  \not\subset
\cB_{\ell+1}$. That is, there exists $v \in \pR_{\ell+1}$ such that
$v \not\in \cB_\ell$. By definition of algorithm, we have
$$R_{\ell+1} = \bB(u_\ell, Q_\ell ) - (\cR_{\ell} \cup \cB_\ell).$$
Therefore,
$$\pR_{\ell+1} \subset (\bB(u_\ell, Q_\ell +1) - \bB(u_\ell, Q_\ell ))\cup \cR_\ell \cup \cB_\ell.$$
 Again, by definition of the algorithm we have
$$B_{\ell+1} = \bB(u_\ell, Q_\ell+1) - \bB(u_\ell, Q_\ell ) - \cR_\ell - \cB_\ell.$$
Therefore, $v \in B_{\ell+1}$ or $v \in \cR_\ell \cup \cB_\ell$.
Recall that by definition of algorithm $\cB_\ell \cap \cR_\ell =
\emptyset$. Since we have assumed that $v \notin \cB_{\ell+1}$, it
must be that $v \in \cR_\ell$. That is, there exists $\ell' \leq
\ell$ such that $v \in R_{\ell'}$.  Now since $v \in \pR_{\ell+1}$
by assumption, it must be that there exists $v' \in R_{\ell+1}$ such
that $\bdg(v,v') = 1$. Since by definition $R_{\ell+1} \cap
R_{\ell'} = \emptyset$, we have $v' \in \pR_{\ell'}$. By induction
hypothesis, this implies that $v' \in \cB_{\ell'} \subset \cB_\ell$.
That is, $\cB_\ell \cap R_{\ell+1} \neq \emptyset$, which is a
contradiction to the definition of our algorithm. That is, our
assumption that $\pR_{\ell +1}  \not\subset \cB_{\ell+1}$ is false.
Thus, we have established the inductive step. This completes the
induction argument and proof of the Claim \ref{claim:one}.
\end{proof}
Now when the algorithm terminates (which must happen within $n$
iterations), say  the output set is $\cB_T$ and $V - \cB_T = \cR_T$
for some $T$. As noted above, $\cR_T$ is a union of disjoint sets
$R_1,\dots, R_T$. We want to show that $R_i, R_j$ are disconnected
for any $1\leq i < j \leq T$ using Claim \ref{claim:one}. Suppose to
the contrary that they are connected. That is, there exists $v \in
R_i$ and $v' \in R_j$ such that $\bdg(v, v') = 1$. Since $R_i \cap
R_j = \emptyset$, it must be that $v' \in \pR_i, v \in \pR_j$. From
Claim \ref{claim:one} and fact that $\cB_t \subset \cB_{t+1}$ for
all $t$, we have that $R_i \cap \cB \neq \emptyset, R_j \cap \cB
\neq \emptyset$. This is contrary to the definition of the
algorithm. Thus, we have established that $R_1,\dots, R_T$ are
disconnected components whose union is $V - \cB_T$. By definition,
each of $R_i \subset \bB(u_{i-1}, K)$. Thus, we have established
that $V - \cB_T$ is made of connected components, each of which is
contained inside balls of radius $K$ with respect to $\bdg$. Since, $G$
has doubling dimension $\rho$, Lemma \ref{lem:poly} implies that
the size of any ball of radius $K$ is at most $(2K)^\rho$. Given
choice of $\beps \leq 1$ and $\rho \geq 1$, we have that $K \geq 2$.
Therefore, $(2K)^\rho \leq K^{2\rho}$.  This
completes the proof of (b) and that of Lemma \ref{lem:decomposition}.
\end{proof}

\subsection{Minor-excluded graphs}

Here we describe a simple and explicit construction of
decomposition for graphs that exclude certain finite
sized graphs as their minor. This scheme is a direct
adapation of a scheme proposed by Klein, Plotkin, Rao \cite{KPR94}
and Rao \cite{R99}. We describe an $(\beps, \Delta)$
node-decomposition scheme. Later, we describe how it can
be modified to obtain $(\beps, \Delta)$ edge-decomposition.

Suppose, we are given graph $G$ that excludes graph $K_{r,r}$
as minor. Recall that if a graph excludes some graph $G_r$
of $r$ nodes as its minor then it excludes $K_{r,r}$ as its
minor as well. In what follows and the rest of the paper, we will
always assume $r$ to be some finite number that does not
scale with $n$ (the number of nodes in $G$).
The following algorithm for generating node-decomposition
uses parameter $\bDelta$.  Later we shall relate the parameter $\bDelta$ to the
decomposition property of the output.

\vspace{.05in} \noindent \DCv($G,r,\bDelta$)\ \vspace{.05in} \hrule
\vspace{.1in}
\begin{itemize}

\item[(0)] Input is graph $G=(V,E)$ and $r, \bDelta \in {\mathbb N}$. Initially,
$i = 0$, $G_0 = G$, $\cB = \emptyset$.

\item[(1)] For $i=0,\dots,r-1$, do the following.
\begin{itemize}
\item[(a)] Let $S^i_{1},\dots,S^i_{k_i}$ be the connected components
of $G_i$.
\item[(b)] For each $S^i_j, 1\leq j\leq k_i$, pick an arbitrary node
$v_j \in S^i_j$.
\begin{itemize}
\item[$\circ$] Create a breadth-first search tree $\cT^i_j$ rooted at $v_j$ in
$S^i_j$.
\item[$\circ$] Choose a number $L^i_j$ uniformly at random from $\{0,\dots, \bDelta-1\}$.
\item[$\circ$] Let $\cB^i_j$ be the set of nodes at level
$L^i_j, \bDelta + L^i_j, 2\bDelta + L^i_j, \dots$ in $\cT^i_j$.

\item[$\circ$] Update $\cB = \cB \cup_{j=1}^{k_i} \cB^i_j$.
\end{itemize}

\item[(c)] set $i = i+1$.
\end{itemize}

\item[(3)] Output $\cB$ and graph $G' = (V, E \backslash \cB)$.
\end{itemize}

\vspace{.1in}
\hrule
\vspace{.05in}

As stated above, the basic idea is to use the following step recursively
(upto depth $r$ of recursion): in each connected component, say $S$,
choose a node arbitrarily and create a breadth-first search tree, say $\cT$.
Choose a number, say $L$, uniformly at random from $\{0,\dots, \bDelta-1\}$.
Remove (and add to $\cB$) all nodes that are at level $L + k \bDelta, k\geq 0$
in $\cT$. Clearly, the total running time of such an algorithm is
$ O(r(n+|E|))$ for a graph $G=(V,E)$ with $|V|=n$; with possible parallel
implementation across different connected components.

\begin{figure}[htb]
\begin{center}
\begin{psfrags}
\psfrag{G0}{$G_0$}
\psfrag{G1}{$G_1$}
\psfrag{1}[r][r]{$1$}
\psfrag{2}[r][r]{$2$}
\psfrag{3}[r][r]{$3$}
\psfrag{4}[r][r]{$4$}
\psfrag{5}[r][r]{$5$}
\psfrag{6}[r][r]{$6$}
\psfrag{7}[r][r]{$7$}
\psfrag{8}[r][r]{$8$}
\psfrag{9}[r][r]{$9$}
\psfrag{S1}{$S_1$}
\psfrag{S2}{$S_2$}
\psfrag{S3}{$S_3$}
\psfrag{L1}{$L^1_1=1$}
\psfrag{T1}{$\cT^1_1$}
\includegraphics[width=7cm]{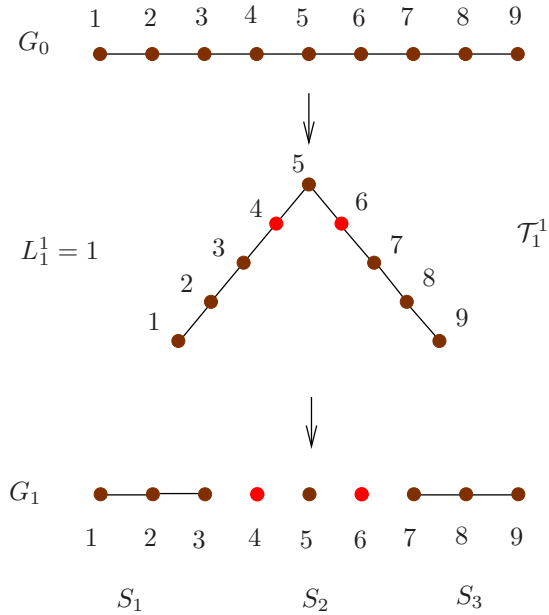}
\end{psfrags}
\end{center}
\caption{The first of two iterations in execution of \DCv($G,2,3$) is shown.}
\label{fig:algo} 
\end{figure}

Figure \ref{fig:algo} explains the algorithm for a line-graph of
$n=9$ nodes, which excludes $K_{2,2}$ as a minor. The example is
about a sample run of \DCv$(G,2,3)$ (Figure \ref{fig:algo} shows
the first iteration of the algorithm).

The following is the result that was in essence proved in \cite{KPR94,R99}. 
\begin{lemma}\label{lem:rao} If $G$ excludes $K_{r,r}$ as a minor. Let $\cB$ be the output of \DCv$(G,r,\bDelta)$.
Then each connected component of $V \backslash \cB$ has diameter of size $O(\bDelta)$.
\end{lemma}
\begin{proof}
This Lemma, for $r = 3$ was proved by Rao in \cite{R99} (Lemma 5 and Corollary 6 of
\cite{R99}).  The result is based on Theorem 4.2 of \cite{KPR94}, which holds for any
$r$. Therefore, the result of Rao naturally extends for any $r$. This completes the
justification of Lemma \ref{lem:rao}. 
\end{proof}
Now using Lemma \ref{lem:rao},  we obtain the following Lemma.
\begin{lemma}\label{lem:dc}
Suppose $G$ excludes $K_{r,r}$ as a minor. Let $d^*$ be maximum
vertex degree of nodes in $G$. Then algorithm
\DCv$(G, r, \bDelta)$ outputs $\cB$ which is
$(r/\bDelta, {d^*}^{O(\bDelta)})$ node-decomposition of $G$.
\end{lemma}
\begin{proof}
Let $R$ be a connected component of $V \backslash \cB$. From 
Lemma \ref{lem:rao}, the diameter of $R$ is $O(\bDelta)$. Since 
$d^*$ is the maximum vertex degree of nodes of $G$,
the number of nodes in $R$ is bounded above by ${d^*}^{O(\bDelta)}$.

To show that $\Pr(v\in \cB)\le r/\bDelta$, consider a vertex 
$v\in V$. If $v\notin \cB$ in the beginning of an iteration $0\le i\le r-1$, then 
it will present in exactly one breadth-first search tree, say $\cT^i_j$. This 
vertex $v$ will be chosen in $\cB^i_j$ only if it is at level $k \bDelta + L^i_j$ 
for some integer $k\ge 0$. The probability of this event is at most 
$1/\bDelta$ since $L^i_j$ is chosen uniformly at random from 
$\{0,1\ldots, \bDelta-1\}$. By union bound, it follows that the 
probability that a vertex  is chosen to be in $\cB$ in any of the 
$r$ iterations is at most $r/\bDelta$. This completes the proof of 
Lemma \ref{lem:dc}.
\end{proof}
It is known that Planar graph excludes $K_{3,3}$ as a
minor. Hence, Lemma \ref{lem:dc} implies the following.
\begin{corollary}\label{cor:planar}
Given a planar graph $G$ with maximum vertex degree $d^*$, then
the algorithm \DCv$(G,3,\bDelta)$ produces
$(3/\bDelta, {d^*}^{O(\bDelta)})$ node-decomposition for any $\bDelta \geq 1$.
\end{corollary}

We describe slight modification of \DCv ~ to obtain algorithm
that produces edge-decomposition as follows. Note that the only
change compared to \DCv ~is the selection of edges rather than
vertices to create the decomposition.

\vspace{.05in} \noindent \DCe($G,r,\bDelta$)\ \vspace{.05in} \hrule
\vspace{.1in}
\begin{itemize}

\item[(0)] Input is graph $G=(V,E)$ and $r, \bDelta \in {\mathbb N}$. Initially,
$i = 0$, $G_0 = G$, $\cB = \emptyset$.

\item[(1)] For $i=0,\dots,r-1$, do the following.
\begin{itemize}
\item[(a)] Let $S^i_{1},\dots,S^i_{k_i}$ be the connected components
of $G_i$.
\item[(b)] For each $S^i_j, 1\leq j\leq k_i$, pick an arbitrary node
$v_j \in S^i_j$.
\begin{itemize}
\item[$\circ$] Create a breadth-first search tree $\cT^i_j$ rooted at $v_j$ in
$S^i_j$.
\item[$\circ$] Choose a number $L^i_j$ uniformly at random from $\{0,\dots, \bDelta-1\}$.
\item[$\circ$] Let $\cB^i_j$ be the set of edges at level
$L^i_j, \bDelta + L^i_j, 2\bDelta + L^i_j, \dots$ in $\cT^i_j$.

\item[$\circ$] Update $\cB = \cB \cup_{j=1}^{k_i} \cB^i_j$.
\end{itemize}

\item[(c)] set $i = i+1$.
\end{itemize}

\item[(3)] Output $\cB$ and graph $G' = (V, E \backslash \cB)$.
\end{itemize}

\vspace{.1in}
\hrule
\vspace{.05in}

\begin{lemma}\label{lem:dce}
Suppose $G$ excludes $K_{r,r}$ as a minor. Let $d^*$ be maximum
vertex degree of nodes in $G$. Then algorithm
\DCe$(G, r, \bDelta)$ outputs $\cB$ which is
$(r/\bDelta, {d^*}^{O(\bDelta)})$ edge-decomposition of $G$.
\end{lemma}
\begin{proof}
Let $G^*$ be a graph that is obtained from $G$ by adding center 
vertex to each edge of $G$. It is easy to see that if $G$ 
excludes $K_{r,r}$ as minor then so does $G^*$.

Now the algorithm \DCe($G,r,\bDelta$) can be viewed as 
executing \DCv($G^*,r,2\bDelta$-1)  with modification that
the random numbers $L^i_j$s are chosen uniformly at random 
from $\{1,3,5,\ldots 2\bDelta-1\}$ instead of the whole support
$\{1,2,\dots, 2\bDelta-1\}$. To prove Lemma \ref{lem:dce}, we need
to show that: (a) each edge is part of the output set $\cB$ with
probability at most $r/\bDelta$, and (b) each of the connected 
component of $V \backslash \cB$ is at most ${d^*}^{O(\bDelta)}$.

The (a) follows from exactly the same arguments as those used in
Lemma \ref{lem:dc}. For (b), consider the following. 
The Lemma \ref{lem:rao} implies that if the algorithm was executed
with the random numbers $L^i_j$s being chosen from $\{1,2,\dots, 2\bDelta-1\}$,
then the desired result follows with probability $1$. It is easy to see
that under the execution of the algorithm with these choices for random
numbers, with strictly positive probability (independent of $n$) all the
$L^i_j$s are chosen only from the odd numbers, i.e. $\{1,3,5,\ldots 2\bDelta-1\}$.
Therefore, it must be that when we restrict the choice of numbers to
these odd numbers, the algorithm must produce the desired result. 
This completes the proof of  Lemma \ref{lem:dce}.
\end{proof}

\begin{figure}[htb]
\begin{center}
\begin{psfrags}
\psfrag{G0}{$G_0$}
\psfrag{G1}{$G_1$}
\psfrag{1}[r][r]{$1$}
\psfrag{2}[r][r]{$2$}
\psfrag{3}[r][r]{$3$}
\psfrag{4}[r][r]{$4$}
\psfrag{5}[r][r]{$5$}
\psfrag{6}[r][r]{$6$}
\psfrag{7}[r][r]{$7$}
\psfrag{8}[r][r]{$8$}
\psfrag{9}[r][r]{$9$}
\psfrag{S1}{$S_1$}
\psfrag{S2}{$S_2$}
\psfrag{S3}{$S_3$}
\psfrag{L1}{$L^1_1=1$}
\psfrag{T1}{$\cT^1_1$}
\includegraphics[width=7cm]{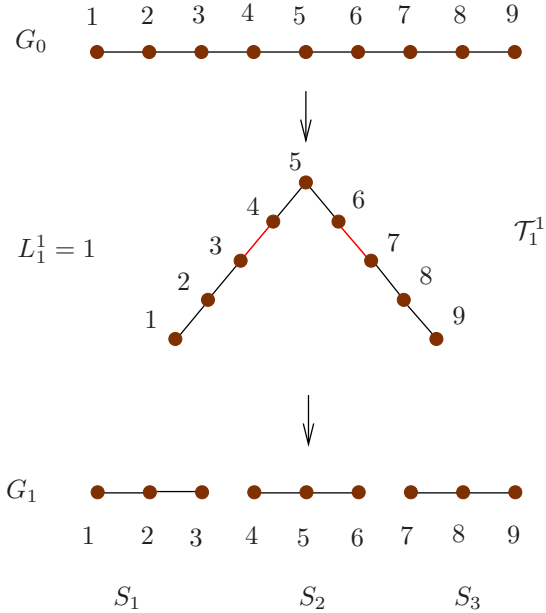}
\end{psfrags}
\end{center}
\caption{The first of two iterations in execution of \DCe($G,2,3$) is shown.}
\label{fig:algo1} 
\end{figure}

Figure \ref{fig:algo1} explains the algorithm for a line-graph of
$n=9$ nodes, which excludes $K_{2,2}$ as a minor. The example is
about a sample run of \DCe$(G,2,3)$ (Figure \ref{fig:algo1} shows
the first iteration of the algorithm).

\section{Approximate $\log Z$}\label{sec:logp}

Here, we describe algorithm for approximate computation of $\log Z$
for any graph $G$. The algorithm uses an edge-decomposition algorithm
as a sub-routine.  Our algorithm provides  provable upper and lower bound
on $\log Z$ for any graph $G$. In order to obtain tight approximation guarantee,
we will use specific graph structures as in low doubling dimension and
minor-excluded graph.

\subsection{Algorithm}

In what follows, we use term \DEC~for a generic
edge-decomposition algorithm.  The approximation guarantee of
the output of the algorithm and its computation time depend
on the property of \DEC. For graph with low doubling dimension,
we use algorithm \DD (over the edge graph) and for graph
that excludes $K_{r,r}$ as minor for some $r$, we use algorithm \DCe.

\vspace{.1in} \noindent \LOGP($G$) \ \vspace{.1in} \hrule
\begin{itemize}


\item[(1)] Use \DEC($G$) to obtain $\cB \subset E$ such that
\begin{itemize}
\item[(a)] $G' = (V, E \backslash \cB)$ is made of connected
components $S_1,\dots, S_K$.
\end{itemize}

\item[(2)] For each connected component $S_j, 1\leq j\leq K$, do the
following:
\begin{itemize}
\item[(a)] Compute partition function $Z_j$ restricted to  $S_j$ by
dynamic programming (or exhaustive computation).
\end{itemize}

\item[(3)] Let $\psi_{ij}^L = \min_{(x,x') \in \Sg^2} \psi_{ij}(x,x')$,
$\psi_{ij}^U = \max_{(x,x') \in \Sg^2} \psi_{ij}(x,x')$. Then
\vspace{-.1in}$$ \log \hZL = \sum_{j=1}^K \log Z_j + \sum_{(i,j) \in
\cB} \psi_{ij}^L; $$
$$ \log \hZU = \sum_{j=1}^K \log Z_j + \sum_{(i,j)
\in \cB} \psi_{ij}^U.$$ \vspace{-.1in}
\item[(4)] Output: lower bound $\log \hZL$ and upper bound $\log \hZU$.

\end{itemize}

\vspace{.1in} \hrule \vspace{.1in}

In words,  \LOGP($G$) produces upper
and lower bound on $\log Z$ of MRF $G$ as follows: decompose
graph $G$ into (small) components $S_1,\dots,S_K$
by removing (few) edges $\cB \subset E$ using \DEC($G$).  Compute
exact log-partition function in each of the components. To
produce bounds $\log \hZL, \log \hZU$ take the summation
of thus computed component-wise log-partition function
along with minimal and maximal effect of edges from
$\cB$.

\subsection{Analysis of \LOGP: General $G$}

Here, we analyze performance of \LOGP~ for any $G$. Later,
we will use property of the specific graph structure to obtain
sharper approximation guarantees.

\begin{theorem}\label{lem:main1}
Given a pair-wise MRF $G$, the \LOGP~ produces
$\log \hZL, \log \hZU$ such that
$$ \log \hZL \leq \log Z \leq \log \hZU, $$
$$ \log \hZU - \log \hZL = \sum_{(i,j) \in \cB} \lf({\psi_{ij}^U}-{\psi_{ij}^L}\rf).$$
It takes $O\lf(|E|{\Sgl}^{|S^*|}\rf) + T_{\text{\DEC}}$ time to
produce this estimate, where $|S^*| = \max_{j=1}^K |S_j|$ with
\DEC~producing decomposition of $G$ into $S_1,\dots,S_K$ in time
$T_{\text{\DEC}}$.
\end{theorem}
\begin{proof}
First, we prove properties of $\log \hZL, \log \hZU$ as follows:
\begin{eqnarray}
 \log \hZL 
&\stackrel{(o)}{=} & \sum_{j=1}^K \log Z_j + \sum_{(i,j) \in \cB}  \psi_{ij}^L \nonumber \\
          & \stackrel{(a)}{=} & \log \lf[\sum_{\bx \in \Sg^n} \exp\lf(\sum_{i \in V} \phi_i(x_i) \rf. \rf. \nonumber \\
          & & ~  \lf.\lf.   + \sum_{(i,j) \in E\backslash \cB} \psi_{ij}(x_i,x_j) + \sum_{(i,j)\in \cB} \psi_{ij}^L \rf) \rf] \nonumber \\
& \stackrel{(b)}{\leq} & \log \lf[\sum_{\bx \in \Sg^n} \exp\lf(\sum_{i \in V} \phi_i(x_i)  \rf. \rf. \nonumber \\
& & ~ \lf.\lf.+ \sum_{(i,j) \in E\backslash \cB} \psi_{ij}(x_i,x_j) + \sum_{(i,j)\in \cB} \psi_{ij}(x_i,x_j) \rf) \rf] \nonumber \\
& = &  \log Z \nonumber \\
& \stackrel{(c)}{\leq} & \log \lf[\sum_{\bx \in \Sg^n} \exp\lf(\sum_{i \in V} \phi_i(x_i) \rf. \rf.  \nonumber \\
& & ~ \lf. \lf. + \sum_{(i,j) \in E\backslash \cB} \psi_{ij}(x_i,x_j) + \sum_{(i,j)\in \cB} \psi^U_{ij} \rf) \rf]  \nonumber \\
& \stackrel{(d)}{=} & \sum_{j=1}^K \log Z_j + \sum_{(i,j) \in \cB}  \psi_{ij}^U \nonumber \\
& = & \log \hZU. \nonumber
 \end{eqnarray}
We justify (a)-(d) as follows: (a) holds because by removal of edges $\cB$, the
$G$ decomposes into disjoint connected components $S_1,\dots, S_K$; (b) holds
because of the definition of $\psi_{ij}^L$; (c) holds by definition $\psi_{ij}^U$
and (d) holds for a similar reason as (a). The claim about difference
$\log \hZU - \log \hZL$ in the statement of Theorem \ref{lem:main1} follows
directly from definitions (i.e. subtract RHS (o) from (d)). This completes
proof of claimed relation between bounds $\log \hZL, \log \hZU$.

For running time analysis, note that \LOGP~ performs two main tasks:
(i) Decomposing $G$ using \DEC~algorithm, which by definition take
$T_{\text{\DEC}}$ time. (ii) Computing $Z_j$ for each component
$S_j$ through exhaustive computation, which takes $O(|E_j|
\Sgl^{|S_j|})$ time (where $E_j$ are edges between nodes of
$S_j$) and producing $\log \hZL, \log \hZU$ takes
addition $|E|$ operations at the most. Now, the maximum size
among these components is $|S^*|$. Further, the $\cup_j E_j \subset E$.
Therefore, we obtain that the total  running time for this task is 
$O(|E| \Sgl^{|S^*|})$. Putting (i) and (ii) together, we obtain the 
desired bound. This completes the proof of Theorem \ref{lem:main1}.
\end{proof}

\subsection{Some preliminaries}

Before stating precise approximation bound of \LOGP~ algorithm for
graphs with low doubling dimension and graphs that exclude minors,
we state two useful Lemmas about $\log Z$ for any graph.

\begin{lemma}\label{lem:main1a}
If $G$ has maximum vertex degree $d^*$ then,
$$ \log Z \geq
\frac{1}{d^*+1} \lf[\sum_{(i,j) \in E} {\psi_{ij}^U}-{\psi_{ij}^L}
\rf].$$
\end{lemma}
\begin{proof}
Assign weight
$w_{ij} = \psi_{ij}^U - \psi_{ij}^L$ to an edge $(i,j) \in E$. Since
graph has maximum vertex degree $d^*$, by Vizing's theorem there
exists an edge-coloring of the graph using at most $d^*+1$ colors.
Edges with the same color form a matching of the $G$. A standard
application of Pigeon-hole's principle implies that there is a color
with weight at least $\frac{1}{d^*+1} (\sum_{(i,j) \in E} w_{ij})$.
Let $M \subset E$ denote these set of edges. That is,
$$ \sum_{(i,j) \in M} (\psi_{ij}^U - \psi_{ij}^L) \geq \frac{1}{d^*+1}\lf(\sum_{(i,j) \in E} (\psi_{ij}^U - \psi_{ij}^L)\rf).$$

Now, consider a $Q \subset \Sg^n$ of size $2^{|M|}$ created as follows.
For $(i,j) \in M$ let
$(x^U_i, x^U_j) \in \arg\max_{(x,x') \in \Sg^2} \psi_{ij}(x,x')$. For
each $i \in V$, choose $x^L_i \in \Sg$ arbitrarily. Then,
\beq
Q & = & \{ \bx \in \Sg^n : \forall~(i,j) \in M, ~(x_i,x_j) = (x^U_i,x^U_j) ~\mbox{or}~\nonumber \\
 & ~ & ~ (x^L_i,x^L_j); \mbox{for~ all other ~$i \in V$},~x_i = x^L_i \}. \nonumber
\eeq
Note that we have used the fact that
$M$ is a matching for $Q$ to be well-defined.

By definition $\phi_i, \psi_{ij}$ are non-negative function (hence,
their exponents are at least $1$). Using this property, we have
the following:
\begin{eqnarray}
Z & \geq &
  \lf[\sum_{\bx \in Q} \exp\lf(\sum_{i \in V} \phi_i(x_i) + \sum_{(i,j)\in E}\psi_{ij}(x_i,x_j)\rf)\rf] \nonumber \\
  & \stackrel{(o)}{\geq} & \lf[\sum_{\bx \in Q} \exp\lf( \sum_{(i,j)\in M}\psi_{ij}(x_i,x_j)\rf)\rf] \nonumber \\
  & \stackrel{(a)}{\geq } & 2^{|M|} \prod_{(i,j) \in M} \frac{\exp(\psi_{ij}^L) + \exp(\psi^U_{ij})}{2} \nonumber 
\end{eqnarray}
\begin{eqnarray}
{\color{white} Z}  & \stackrel{(b)}{=} & \prod_{(i,j) \in M} (1 + \exp(\psi_{ij}^U - \psi_{ij}^L)) \exp(\psi_{ij}^L)~~~~~~~~~~~~ \nonumber \\
  & \stackrel{(c)}{\geq} & \prod_{(i,j) \in M} \exp(\psi_{ij}^U - \psi_{ij}^L) \nonumber \\
  & = & \exp\lf(\sum_{(i,j)\in M} \psi_{ij}^U- \psi_{ij}^L\rf). \label{e:1}
\end{eqnarray}
Justification of (o)-(c): (o) follows since $\psi_{ij}, \phi_i$ are
non-negative functions. (a) consider the following probabilistic
experiment: assign $(x_i,x_j)$ for each $(i,j) \in M$ equal to
$(x^U_i, x^U_j)$ or $(x^L_i, x^L_j)$ with probability $1/2$ each.
Under this experiment, the expected value of the $\exp(\sum_{(i,j)
\in M} \psi_{ij}(x_i,x_j))$, which is $\prod_{(i,j) \in M}
\frac{\exp(\psi_{ij}(x_i^L, x_j^L)) +
\exp(\psi_{ij}(x_i^U,x_j^U))}{2}$, is equal to $2^{-|M|} [\sum_{\bx
\in Q}\exp(\sum_{(i,j) \in M} \psi_{ij}(x_i,x_j))]$. Now, use the
fact that $\psi_{ij}(x_i^L, x_j^L) \geq \psi_{ij}^L$. (b) follows
from simple algebra and (c) follows by using non-negativity of
function $\psi_{ij}$. Therefore,
\begin{eqnarray}
\log Z & \geq & \sum_{(i,j) \in M} \lf(\psi_{ij}^U - \psi_{ij}^L\rf) \nonumber \\
      & \geq & \frac{1}{d^*+1} \lf(\sum_{(i,j) \in E} \lf(\psi_{ij}^U - \psi_{ij}^L\rf)\rf),
\end{eqnarray}
using fact about weight of $M$. This completes the proof of Lemma
\ref{lem:main1a}.
\end{proof}

\begin{lemma}\label{lem:main1b}
If $G$ has maximum vertex degree $d^*$ and the \DEC($G$) produces
$\cB$ that is $(\beps, \Delta)$ edge-decomposition, then
$$ \Ex\lf[\log \hZU - \log \hZL\rf]  \leq  \beps (d^*+1) \log Z,$$ w.r.t. the
randomness in $\cB$, and \LOGP ~takes time $O(n d^* \Sgl^{\Delta}) +
T_{\text{\DEC}}$.
\end{lemma}
\begin{proof}
From Theorem  \ref{lem:main1}, Lemma \ref{lem:main1a} and definition of $(\beps,
\Delta)$ edge-decomposition, we have the following.
\begin{eqnarray}
\Ex\lf[\log \hZU - \log \hZL\rf] & \leq & \Ex\lf[\sum_{(i,j)\in \cB} (\psi_{ij}^U-\psi_{ij}^L)\rf] \nonumber \\
& = & \sum_{(i,j) \in E} \Pr((i,j) \in \cB) (\psi_{ij}^U-\psi_{ij}^L) \nonumber \\
& \leq & \beps \lf[\sum_{(i,j) \in E} (\psi_{ij}^U-\psi_{ij}^L)\rf] \nonumber \\
& \leq & \beps (d^*+1) \log Z. \nonumber
\end{eqnarray}
Now to estimate the running time, note that under $(\beps, \Delta)$
decomposition $\cB$, with probability $1$ the $G'=(V, E\backslash \cB)$
is divided into connected components with  at most $\Delta$ nodes.
Therefore, the running time bound of Theorem \ref{lem:main1} implies
the desired result.
\end{proof}

\subsection{Analysis of \LOGP: Low doubling-dimension $G$}

Here we interpret result obtained in Theorem \ref{lem:main1} and
Lemma \ref{lem:main1b}, for $G$ that has low doubling-dimension
and uses decomposition scheme \DD.

\begin{theorem}\label{thm:main0}
Let MRF graph $G$ of $n$ nodes with doubling dimension $\rho$ be given.
Consider any $\beps \in (0,1)$. Define $\bdelta = \beps 2^{-\rho - 3}$.  
Then \LOGP~ using \DD($\bdelta, K(\bdelta, \rho)$)  produces 
bounds  $\log \hZL, \log \hZU$  such that
$$\Ex\lf[\log \hZU - \log \hZL\rf] \leq \beps \log Z.$$
The algorithm takes $O(n 2^\rho C_0(\beps, \rho))$ time to obtain
the estimate, where $C_0(\beps, \rho) = |\Sigma|^{K(\bdelta, \rho)^{2\rho}}$. 
Further, if $\rho (\rho + \log 1/\beps) = o(\log \log n)$ then the
algorithm takes $o(n^{1+\delta})$ amount of time for any $\delta > 0$.
\end{theorem}
\begin{proof}
The Lemma \ref{lem:decomposition}, Lemma \ref{lem:main1b} 
and Theorem \ref{lem:main1} implies the following bound:
\begin{eqnarray}
\Ex\lf[\log \hZU - \log \hZL\rf] & \leq & \beps 2^{-\rho-2} (d^*+1) \log Z \nonumber \\
      & \leq & \beps \log Z.
\end{eqnarray}
Now for graph with doubling dimension $\rho$,  $|E| = n 2^\rho$. Under
the decomposition algorithm with parameter $\bdelta$ and $K(\bdelta, \rho)$,
the number of nodes in any component is at most $K(\bdelta,\rho)^{2\rho}$. 
Therefore, by Lemma \ref{lem:decomposition} the desired bound on 
running time follows. 

Now, consider when condition $\rho (\rho + \log 1/\beps) = o(\log \log n)$. 
Given $\bdelta = \beps 2^{-\rho - 3}$,
\beq
\rho \log \rho/\bdelta & = & \rho (\log \rho + \rho + 3 + \log 1/\beps) \nonumber \\
  & = & \Theta(\rho^2 + \rho \log 1/\beps) \nonumber \\
  & = & o(\log \log n),
\eeq
from the above described hypothesis of the Theorem.  Now, 
\DD($\bdelta, K(\bdelta, \rho)$) produces $(\beps 2^{-\rho -2}, O(\log^{1/L} n))$ 
edge-decomposition from Corollary \ref{cor:decomposition}.  We select 
$L = 2$. Given this and above arguments, we have that the running time of 
the algorithm is $o(n^{1+\delta}))$ for any $\delta > 0$. This completes the proof
of Theorem \ref{thm:main0}.
\end{proof}

\subsection{Analysis of \LOGP: Minor-excluded $G$}

We apply Theorem \ref{lem:main1} and Lemma \ref{lem:main1b} for
minor-excluded graphs when the \DEC~procedure is essentially
the \DCe.  We obtain the following precise result.

\begin{theorem}\label{thm:main1}
Let MRF graph $G$ of $n$ nodes exclude $K_{r,r}$ as its minor. Let
$d^*$ be the maximum vertex degree in $G$. Given $\beps > 0$,
use \LOGP~algorithm with \DCe($G,r,\bDelta$) where
$\bDelta = \lceil \frac{r (d^*+1)}{\beps}\rceil$.
Then,
$$ \log \hZL \leq \log Z \leq \log \hZU; ~\mbox{and}~$$
$$ \Ex\lf[\log \hZU - \log \hZL\rf] \leq \beps \log Z.$$
Further, algorithm takes $(n C(d^*,\Sgl,\beps))$, where constant
$C(d^*,\Sgl,\beps) = d^* \Sgl^{{d^*}^{O(\Lambda)}}.$ Therefore,
if $\beps^{-1} d^*\log d^*  = o(\log \log n)$, then the algorithm takes $o(n^{1+\delta})$ 
steps for arbitrary $\delta > 0$.
\end{theorem}
\begin{proof}
From Lemma \ref{lem:dce} about the \DCe~ algorithm, we have
that with choice of $\Lambda = \lceil \frac{r (d^*+1)}{\beps}\rceil$,
the algorithm produces $(\beps, \Delta)$ edge-decomposition
where $\Delta ={d^*}^{O(\Lambda)}$. Since it is an $(\beps, \Delta)$
edge-decomposition, the upper bound and the lower bound, $\log \hZU, \log \hZL$,
for the value produced by the algorithm are  within $(1\pm \beps)\log Z$ by Lemma \ref{lem:main1b}.

Now, by Lemma \ref{lem:main1b} the running time of the algorithm is
$O(n d^* \Sgl^\Delta) + T_{\text{\DEC}}$. As discussed earlier in Lemma
\ref{lem:dce}, the algorithm \DCe ~takes $O(r|E|) = O(nrd^*)$ operations.
That is, $T_{\text{\DEC}} = O(nrd^*)$.  Now, $\Delta = {d^*}^{O(\Lambda)}$
and $\Lambda \le r (d^*+1)/\beps +1$. Therefore, the first term of
the computation time bound is bounded above by
$$O\lf(n d^* \Sgl^{{d^*}^{O( r d^*/\beps)}}\rf).$$
Now, we will establish that the above term is $O(n^2)$ under the
hypothesis $\beps^{-1} d^*\log d^* = o(\log \log n)$. The hypothesis
implies that (since $r$ a constant, not scaling with $n$):
$$ \Lambda \log d^* = o(\log \log n).$$
That is, for any finite $L$ (say, $L=2$) we have that
$$ \Delta = O(\log^{1/L} n).$$
This in turn implies that, for finite $\Sgl$ we have
$$ \Sgl^{\Delta} = o(n^{\delta/2}),$$
for any $\delta > 0$.  Since $d^* = o(\log \log n) = O(n^{\delta/2})$.
Therefore, it follows that 
$$O\lf(n d^* \Sgl^{\Delta}\rf) = o(n^{1+\delta}).$$
This completes the proof of Theorem \ref{thm:main1}.
\end{proof}

\section{Approximate MAP}\label{sec:map}

Now, we describe algorithm to compute MAP
approximately. It is very similar to the \LOGP~algorithm: given $G$,
decompose it into (small) components $S_1,\dots,S_K$ by removing
(few) edges $\cB \subset E$.  Then, compute an approximate MAP
assignment by computing exact MAP restricted to the components. As
in \LOGP, the computation time and performance of the algorithm
depends on property of decomposition scheme. We describe algorithm
for any graph $G$; which will be specialized for graph with low
doubling dimension and graph that exclude minor by using the
appropriate edge-decomposition schemes.

\vspace{.1in} \noindent \MAP($G$) \ \vspace{.05in} \hrule
\vspace{.1in}
\begin{itemize}
\vspace{.1in}
\item[(0)] Input is MRF $G=(V,E)$ with $\phi_i(\cdot), i \in V$,
$\psi_{ij}(\cdot,\cdot), (i,j) \in E$.
\item[(1)] Use \DEC($G$) to obtain $\cB \subset E$ such that
\begin{itemize}
\item[(a)] $G' = (V, E \backslash \cB)$ is made of connected
components $S_1,\dots, S_K$.
\end{itemize}

\item[(2)] For each connected component $S_j, 1\leq j\leq K$, do the
following:
\begin{itemize}
\item[(a)] Through dynamic programming (or exhaustive computation)
find exact MAP $\bx^{*,j}$ for component $S_j$, where
$\bx^{*,j}=(x^{*,j}_i)_{i\in S_j}$.
\end{itemize}

\item[(3)] Produce output $\widehat{\bx^*}$, which is obtained by
assigning values to nodes using $\bx^{*,j}, 1\leq j\leq K$.

\end{itemize}

\vspace{.02in} \hrule \vspace{.05in}

\subsection{Analysis of \MAP: General $G$}

Here, we analyze performance of \MAP~ for any $G$. Later, we
will specialize our analysis for graph with low doubling dimension
and minor excluded graphs.

\begin{theorem}\label{lem:main2}
Given an MRF $G$ described by (\ref{markovicity}), the \MAP~
algorithm produces outputs $\widehat{\bx^*}$ such that:
$$ \cH(\bx^*) - \sum_{(i,j) \in \cB} \lf({\psi_{ij}^U}-{\psi_{ij}^L}\rf) ~ \leq
\cH(\widehat{\bx^*}) \leq \cH(\bx^*).$$
The algorithm takes $O\lf(|E|K {\Sgl}^{|S^*|}\rf) + T_{\text{\DEC}}$ time to
produce this estimate, where $|S^*| = \max_{j=1}^K |S_j|$ with \DEC~producing
decomposition of $G$ into $S_1,\dots,S_K$ in time $T_{\text{\DEC}}$.
\end{theorem}
\begin{proof}
By definition of MAP $\bx^*$, we have $\cH(\widehat{\bx^*}) \leq \cH(\bx^*)$. Now,
consider the following.
\begin{eqnarray}
\cH({\bx^*}) & = & \max_{\bx \in \Sg^n} \lf[\sum_{i \in V} \phi_i(x_i) + \sum_{(i,j) \in E} \psi_{ij}(x_i,x_j)\rf] \nonumber \\
& = &  \max_{\bx \in \Sg^n} \lf[\sum_{i \in V} \phi_i(x_i) + \sum_{(i,j) \in E\backslash \cB} \psi_{ij}(x_i,x_j) \rf. \nonumber \\
& ~ & ~~~ \lf. + \sum_{(i,j)\in \cB} \psi_{ij}(x_i, x_j)\rf] \nonumber \\
 & \stackrel{(a)}{\leq} & \max_{\bx \in \Sg^n} \lf[\sum_{i \in V} \phi_i(x_i) + \sum_{(i,j) \in E\backslash \cB} \psi_{ij}(x_i,x_j) \rf. \nonumber \\
 & ~ & ~~~ \lf. + \sum_{(i,j)\in \cB} \psi^U_{ij}\rf] \nonumber \\
& \stackrel{(b)}{=} & \sum_{j=1}^K  \lf[\max_{\bx^j \in \Sg^{|S_j|}} \cH(\bx^j)\rf] + \lf[\sum_{(i,j)\in \cB} \psi_{ij}^U\rf] \nonumber \\
& \stackrel{(c)}{=} & \sum_{j=1}^K \cH(\bx^{*,j}) + \lf[\sum_{(i,j)\in \cB} \psi_{ij}^U \rf] \nonumber \\
& \stackrel{(d)}{\leq} & \cH(\widehat{\bx^*}) + \lf[\sum_{(i,j)\in \cB} \psi_{ij}^U-\psi_{ij}^L\rf].
\end{eqnarray}
We justify (a)-(d) as follows: (a) holds because for each edge $(i,j) \in \cB$, we
have replaced its effect by maximal value $\psi_{ij}^U$; (b) holds
because by placing constant value $\psi_{ij}^U$ over $(i,j) \in \cB$, the
maximization over $G$ decomposes into maximization over the connected
components of $G'=(V, E\backslash \cB)$; (c) holds by definition of $\bx^{*,j}$
and (d) holds because when we obtain global assignment $\widehat{\bx^*}$ from
$\bx^{*,j}, 1\leq j\leq K$ and compute its global value, the additional terms
get added for each $(i,j) \in \cB$ which add at least $\psi_{ij}^L$ amount.

The running time analysis of \MAP~ is exactly the same as that of \LOGP~ in
Theorem \ref{lem:main1}. Hence, we skip the details here.
This completes the proof of Theorem \ref{lem:main2}.

\end{proof}

\subsection{Some preliminaries}

This section presents some results about the property of
MAP solution that will be useful in obtaining tight approximation
guarantees later. First, consider the following.

\begin{lemma}\label{lem:main2a}
If $G$ has maximum vertex degree $d^*$, then
\beq
\cH(\bx^*) & \geq &  \frac{1}{d^*+1} \lf[\sum_{(i,j) \in E} {\psi_{ij}^U}\rf] ~\nonumber \\
& \geq & \frac{1}{d^*+1} \lf[\sum_{(i,j) \in E} {\psi_{ij}^U}-{\psi_{ij}^L} \rf].
\eeq
\end{lemma}
\begin{proof}
Assign weight $w_{ij} = \psi_{ij}^U$ to an edge $(i,j) \in E$. Using
argument of Lemma \ref{lem:main1a}, we obtain that there exists a
matching $M \subset E$ such that
$$ \sum_{(i,j) \in M} \psi_{ij}^U \geq \frac{1}{d^*+1}\lf(\sum_{(i,j) \in E} \psi_{ij}^U\rf).$$
Now, consider an assignment $\bx^M$ as follows: for each $(i,j) \in M$
set $(x_i^M, x_j^M) = \arg\max_{(x,x') \in \Sg^2} \psi_{ij}(x,x')$; for
remaining $i \in V$, set $x_i^M$ to some value in $\Sigma$ arbitrarily.
Note that for above assignment to be possible, we have used
matching property of $M$. Therefore, we have
\begin{eqnarray}
\cH(\bx^M) & = & \sum_{i \in V} \phi_i(x_i^M) + \sum_{(i,j) \in E} \psi_{ij}(x_i^M, x_j^M) \nonumber \\
& = & \sum_{i \in V} \phi_i(x_i^M) + \sum_{(i,j) \in E \backslash M} \psi_{ij}(x_i^M, x_j^M) \nonumber \\
& ~ & ~~ + \sum_{(i,j) \in M} \psi_{ij}(x_i^M, x_j^M) \nonumber \\
& \stackrel{(a)}{\geq} & \sum_{(i,j) \in M} \psi_{ij}(x_i^M, x_j^M) \nonumber \\
& = & \sum_{(i,j) \in M} \psi_{ij}^U \nonumber \\
& \geq & \frac{1}{d^*+1} \lf[\sum_{(i,j)\in E} \psi_{ij}^U\rf].
\end{eqnarray}
Here (a) follows because $\psi_{ij}, \phi_i$ are non-negative valued
functions.  Since $\cH(\bx^*) \geq \cH(\bx^M)$ and $\psi_{ij}^L \geq
0$ for all $(i,j) \in E$, we obtain the Lemma \ref{lem:main2a}.
\end{proof}

\begin{lemma}\label{lem:main2b}
If $G$ has maximum vertex degree $d^*$ and the \DEC($G$)
produces $\cB$ that is $(\beps, \Delta)$ edge-decomposition,
then
$$ \Ex\lf[\cH(\bx^*) - \cH(\widehat{\bx^*}) \rf]  \leq  \beps (d^*+1) \cH(\bx^*),$$
where expectation is w.r.t. the randomness in $\cB$.
Further, \MAP ~takes time $O(n d^* \Sgl^{\Delta}) + T_{\text{\DEC}}$.
\end{lemma}
\begin{proof}
From Theorem \ref{lem:main2}, Lemma \ref{lem:main2a} and definition of $(\beps,
\Delta)$ edge-decomposition, we have the following.
\begin{eqnarray}
\Ex\lf[\cH(\bx^*) - \cH(\widehat{\bx^*}) \rf] & \leq &
\Ex\lf[\sum_{(i,j)\in \cB} (\psi_{ij}^U-\psi_{ij}^L)\rf] \nonumber \\
& = & \sum_{(i,j) \in E} \Pr((i,j) \in \cB) (\psi_{ij}^U-\psi_{ij}^L) \nonumber \\
& \leq & \beps \lf[\sum_{(i,j) \in E} (\psi_{ij}^U-\psi_{ij}^L)\rf] \nonumber \\
& \leq & \beps (d^*+1) \cH(\bx^*).
\end{eqnarray}
The running time bound can be obtained using arguments similar to
those in Lemma \ref{lem:main1b}.
\end{proof}

\subsection{Analysis of \MAP: Low doubling dimension $G$}

Here we interpret result obtained in Theorem \ref{lem:main2} and
Lemma \ref{lem:main2b}, for $G$ that has low doubling-dimension
and uses decomposition scheme \DD.

\begin{theorem}\label{thm:main0x}
Let MRF graph $G$ of $n$ nodes with doubling dimension $\rho$ be given.
Consider any $\beps \in (0,1)$ such that $\rho (\rho + \log 1/\beps) = o(\log \log n)$,
define $\bdelta = \beps 2^{-\rho - 3}$.  Then \MAP~ using
\DD($\bdelta, K(\bdelta, \rho)$)  produces bounds  $\widehat{\bx^*}$  such that
$$\Ex\lf[\cH(\bx^*) - \cH(\widehat{\bx^*}) \rf] \leq \beps \cH(\bx^*).$$
The algorithm takes $O(n 2^\rho C_0(\beps, \rho))$ time to obtain
the estimate, where $C_0(\beps, \rho) = |\Sigma|^{K(\bdelta, \rho)^{2\rho}}$. 
Further, if $\rho (\rho + \log 1/\beps) = o(\log \log n)$ then the
algorithm takes $o(n^{1+\delta})$ amount of time for any $\delta > 0$.
\end{theorem}
\begin{proof}
Theorem \ref{lem:main2}, Lemma \ref{lem:main2b} and Lemma \ref{lem:decomposition}
imply that the output produced
by \MAP ~algorithm is such that
\begin{eqnarray}
\Ex\lf[\cH(\bx^*) - \cH(\widehat{\bx^*})\rf] & \leq & \beps 2^{-\rho-2} (d^*+1) \cH(\bx^*) \nonumber \\
      & \leq & \beps \cH(\bx^*),
\end{eqnarray}
because $d^*+1 \leq 2^{\rho + 2}$ for a graph with doubling dimension $\rho$.
The running time analysis of the algorithm follows exactly the same
arguments as those in the proof of Theorem \ref{thm:main0}. 

\end{proof}

\subsection{Analysis of \MAP: Minor-excluded $G$}

We apply Theorem \ref{lem:main2} and Lemma \ref{lem:main2b} for
minor-excluded graphs when the \DEC~procedure is 
the \DCe.  We obtain the following precise result.

\begin{theorem}\label{thm:main1x}
Let MRF graph $G$ of $n$ nodes exclude $K_{r,r}$ as its minor. Let
$d^*$ be the maximum vertex degree in $G$. Given $\beps > 0$,
use \MAP~algorithm with \DCe($G,r,\bDelta$) where
$\bDelta = \lceil \frac{r (d^*+1)}{\beps}\rceil$.
Then,
$$\Ex\lf[\cH(\bx^*) - \cH(\widehat{\bx^*}) \rf] \leq \beps \cH(\bx^*).$$
Further, algorithm takes $(n C(d^*,\Sgl,\beps))$, where
$C(d^*,\Sgl,\beps) = d^* \Sgl^{{d^*}^{O(\bDelta)}}.$ Therefore,
if $\beps^{-1} d^*\log d^*  = o(\log \log n)$, then the algorithm takes $o(n^{1+\delta})$ steps
for arbitrary $\delta > 0$.
\end{theorem}
\begin{proof}
From Lemma \ref{lem:dce} about the \DCe~ algorithm, we have
that with choice of $\bDelta = \lceil \frac{r (d^*+1)}{\beps}\rceil$,
the algorithm produces $(\beps, \Delta)$ edge-decomposition
where $\Delta = {d^*}^{O(\bDelta)}$. Since its an $(\beps, \Delta)$
edge-decomposition, from Lemma \ref{lem:main2b} it follows that
$$\Ex\lf[\cH(\bx^*) - \cH(\widehat{\bx^*}) \rf] \leq \beps \cH(\bx^*).$$

Now, by Lemma \ref{lem:main2b} the algorithm running time is
$O(n d^* \Sgl^\Delta) + T_{\text{\DEC}}$. As discussed earlier in Lemma
\ref{lem:dce}, the algorithm \DCe ~takes $O(r|E|) = O(nrd^*)$ operations.
That is, $T_{\text{\DEC}} = O(nrd^*)$.  Now, $\Delta = {d^*}^{O(\Lambda)}$
and $\Lambda \leq  r (d^*+1)/\beps +1$. Therefore, the first term of
the computation time bound is bounded above by
$$O\lf(n d^* \Sgl^{{d^*}^{O( r {d^*}/\beps)}}\rf).$$
Now, we will establish that the above term is $O(n^2)$ under the
hypothesis $\beps^{-1} d^*\log d^* = o(\log \log n)$. The hypothesis
implies that (since $r$ a constant, not scaling with $n$):
$$ \Lambda \log d^* = o(\log \log n).$$
That is, for any finite $L$ (say, $L=2$) we have that
$$ \Delta = O(\log^{1/L} n).$$
This in turn implies that, for finite $\Sgl$ we have
$$ \Sgl^{\Delta} = o(n^{\delta/2}),$$
for any $\delta > 0$.  Since $d^* = o(\log \log n) = O(n^{\delta/2})$.
Therefore, it follows that 
$$O\lf(n d^* \Sgl^{\Delta}\rf) = o(n^{1+\delta}).$$
This completes the proof of Theorem \ref{thm:main1x}.
\end{proof}

\section{Message-passing implementation through self-avoiding walk}\label{sec:mp}

The approximate inference algorithms, \LOGP ~and~\MAP ~presented
above are {\em local} in the sense that in order to make computation,
the centralization of the algorithm is limited only up to each connected
component. This section provides a method for designing message-passing
implementation for computing these estimates using the self-avoiding
walk trees. This message passing algorithm is explained for MAP
computation and is restricted to binary MRF. It is worth noting that
any MAP estimation problem over discrete pair-wise exponential
family can be converted into a binary pair-wise MRF with the help
of addition nodes. This is explained in Appendix \ref{ap:binary}.
Thus, in principle, this message passing algorithm can work for
any discrete valued Markov random field represented by a
factor graph.



\subsection{Equivalence: MRF and Self-Avoiding Walk Tree}\label{ssec:1-1}

The first result is about equivalence of max-marginal of a node, say $v$, in
an MRF $G$ and max-marginal of root of self-avoiding walk
tree with respect to $v$. Dror Weitz \cite{DW06} showed such equivalence
in the context of marginal distributions of the nodes. We establish the
result for max-marginal. However, the proof is a direct adaption
of the proof of result by Weitz \cite{DW06}.

\begin{figure}[htb]
\begin{center}
\begin{psfrags}
\psfrag{p12}{$\psi_{12}$} \psfrag{p13}{$\psi_{13}$}
\psfrag{p23}{$\psi_{23}$} \psfrag{p34}{$\psi_{34}$}
\psfrag{1}[r][r]{$1$} \psfrag{2}[r][r]{$2$} \psfrag{3}[r][r]{$3$}
\psfrag{4}[r][r]{$4$} \psfrag{1o}[r][r]{{\color{green} $1$}}
\psfrag{1c}[r][r]{{\color{red} $1$}} \psfrag{tsaw}[r][r]{$\Ts(G,1)$}
\psfrag{tcomp}[l][l]{$\Tc(G,1)$} \psfrag{G}[r][r]{Graph $G$}
\includegraphics[width=8cm]{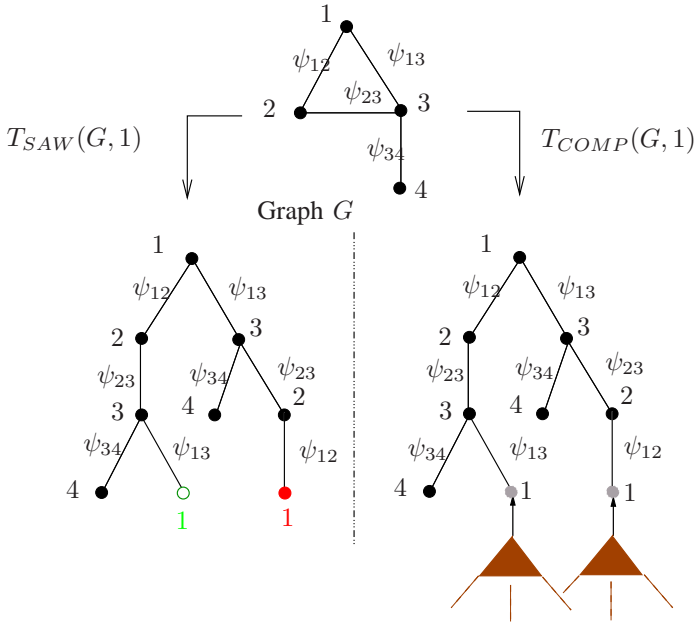}
\end{psfrags}
\end{center}
\caption{A graph $G$ of $4$ nodes with one loop is given. On left,
we have the self-avoiding walk tree of $G$ for node $1$, i.e.
$\Ts(G, 1)$ with green and red being special nodes. On right, we
have computation tree $\Tc(G,1)$ for node $1$'s computation under
Belief Propagation (or Max-Product) algorithm. The grey nodes of
$\Tc(G,1)$ correspond to green and red node of $\Ts(G,1)$ on the
left.} \label{fig:onex} 
\end{figure}

Given binary pair-wise MRF $G$ of $n$ nodes, our interest is
in finding
$$p^*_v(\gamma ) = \max_{{\sigma \in \{0,1\}^n: }{\sigma_v = \gamma}} \Pr(\sigma),
\mbox{~for~} \gamma \in \{0,1\}~\mbox{for~all}~ v.$$

\begin{definition}[Self-Avoiding Walk Tree]\label{self}
Consider graph $G=(V,E)$ of pair-wise binary MRF. For $v \in V$, we
define the self avoiding walk tree $\Ts(G,v)$ as follows. First, for
each $u \in V$, give an ordering of its neighbors $N(u)$. This
ordering can be arbitrary but remains fixed forever. Given this,
$\Ts(G,v)$ is constructed by the breadth first search of nodes of
$G$ starting from $v$ without backtracking. Then stop the
bread-first search along a direction when an already visited vertex
is encountered (but include it in $\Ts(G,v)$ as a leaf). Say one
such leaf be $\hat{w}$ of $\Ts(G,v)$ and let it be a copy of a node
$w$ in $G$. We call such a leaf node of $\Ts(G,v)$ as {\em Marked}.
A marked leaf node is assigned color {\em Red} or {\em Green}
according to the following condition: The leaf $\hat{w}$ is marked
since we encountered node ${w}$ of $G$ twice along our bread-first
search excursion. Let the (directed) path between these two
encounters of $w$ in $G$ be given by $(w, v_1,\dots,v_k, w)$.
Naturally, $v_1, v_k \in N(w)$ in $G$. We mark the leaf node
$\hat{w}$ as {\em Green} if according to the ordering done by node
$w$ in $G$ of its neighbors, if $v_k$ is given smaller number than
that of $v_1$. Else, we mark it as {\em Red}. Let ${\bf V}_v$ and
${\bf E}_v$ denote the set of nodes and vertices of tree $\Ts(G,v)$.
With little abuse of notation, we will call root of $\Ts(G,v)$ as
$v$.
\end{definition}

Given a $\Ts(G,v)$ for a node $v \in V$ in $G$, an MRF is naturally
induced on it as follows: all edges inherit the pair-wise compatibility
function (i.e. $\psi_{\cdot\cdot}(\cdot,\cdot)$) and all
nodes inherit node-potentials (i.e. $\phi_{\cdot}(\cdot)$) from
those of MRF $G$ in a natural manner. The only distinction is the
modification of the node-potential of {\em marked} leaf nodes
of $\Ts(G,v)$ as follows. A marked leaf node, say $\hat{w}$
 of $\Ts(G, v)$ modifies
its potentials as follows: if it is {\em Green} than it sets
$\phi_{\hat{w}}(1) = \phi_w(1), \phi_{\hat{w}}(0)= 0$ but if
it is {\em Red} leaf node then it sets
$\phi_{\hat{w}}(0) = \phi_w(0), \phi_{\hat{w}}(1)= 0$.

\begin{example}[Self-avoiding walk tree] Consider $4$ node binary pair-wise MRF $G$ in
Figure \ref{fig:onex}. Let node $1$ gives number $a$ to node $2$,
number $b$ to node $3$ so that $a > b$. Given this numbering, the
bottom left of Figure \ref{fig:onex} represents $\Ts(G,1)$. The Green
leaf node essentially means that we set its value permanently to
$1$.
\end{example}

With above description, $\Ts(G,v)$ gives rise to a pair-wise binary
MRF. Let $\Qr_{G,v}$ denote the probability distribution induced by
this MRF on boolean cube $\{0,1\}^{|{\bf V}_v|}$. Our interest will
be in the max-marginal for root $v$ or equivalently
$$q^*_v(\gamma) = \max_{\sigma \in \{0,1\}^{|{\bf V}_v|}: \sigma_v =\gamma} \Qr_{G,v}(\sigma), \mbox{~where~}
\gamma \in \{0,1\}.$$

Here we present an equivalence between $p^*_v(\cdot)$ and $q^*_v(\cdot)$. This
is a direct adaptation of result by Weitz \cite{DW06}.
\begin{theorem}\label{thm:two}
Consider any binary pair-wise MRF $G = (V,E)$. For any $v \in V$,
let $p^*_v(\cdot)$ be as defined  above with respect to $\Pr_G$. Let
$\Ts(G,v)$ be the self-avoiding walk tree MRF and let $q^*_v(\cdot)$
be as defined above for root node of $\Ts(G,v)$ with respect to
$\Qr_{G,v}$. Then, \beq \frac{p^*_v(1)}{p^*_v(0)} & = &
\frac{q^*_v(1)}{q^*_v(0)}.\label{eq:1} \eeq Here we allow ratio to
be $0, \infty$.
\end{theorem}
\begin{proof}
The proof follows by induction. As a part of the proof, we will come
across graphs with some {\em fixed} vertices, where a vertex $u$ is
said to be fixed to $0$ (resp. $1$) if $\phi_u(0)>0$ , $\phi_u(1)=0$
(resp. $\phi_u(1)> 0$ , $\phi_u(0)=0$). The induction is on the
number of {\em unfixed} vertices of $G$. We essentially prove the
following, which implies the statement of Lemma: given any pair-wise
MRF on a graph $G$ (with possibly some \fx~vertices), construct
corresponding $\Ts(G,v)$ MRF for some node $v$. If the number of
{\em unfixed} vertex of $G$ is at most $m$, then the (\ref{eq:1})
holds. Next, inductive proof.

\vspace{.05in}
\noindent{\em Initial condition.} Trivially the desired statement
holds for any graph with exactly one {\em unfixed}  vertex,
by definition of MRF, i.e. (\ref{markovicity}). The reason is that
for such a graph, due to all but one node being fixed, the
max-marginal of each node is purely determined by its immediate
neighbors due to Markovian nature of MRF. The immediate
neighborhood of $v$ in $\Ts(G,v)$ and $G$ is the same.

\vspace{.05in} \noindent{\em Hypothesis.} Assume that the statement
is true for any graph with less than or equal to $m \in \bbN$ {\em
unfixed} nodes.

\noindent{\em Induction step.} Without loss of generality, suppose
that our graph of interest, $G$, has $m+1$ {\em unfixed} vertices.
If $v$ is a \fx~vertex, then (\ref{eq:1}) holds trivially.  Let
$v\in V$ be an unfixed vertex of $G$. Then we will show via
inductive hypothesis that
$$\frac{q^*_v(1)}{q^*_v(0)}=\frac{p^*_v(1)}{p^*_v(0)}.$$

Let $d$ be the degree of $v$;  $v_1,v_2,\ldots,v_d$ be the neighbors
of $v$ where the order of neighbors is the same as that used in
definition of $\Ts(G,v)$. Let $T_\ell$ be the $\ell$th subtree of
$\Ts(G,i)$ having $v_\ell$ as its root and $Y(\ell)$ be the binary
pair-wise MRF induced on $T_\ell$ by restriction of $\Ts(G,v)$. Let
$q^{*}_\ell(\sigma)$ be the max-marginal of vertex $v_\ell$ taking
value $\sigma \in \Sg = \{0,1\}$ with respect to $Y(\ell)$. Note
that when $T_\ell$ consists of a single vertex, then
$q^{*}_\ell(\sigma) \propto \phi_{v_\ell}(\sigma)$. Let $\lambda_v
=\frac{\phi_v(1)}{\phi_v(0)}$. Then from definition of  pair-wise
MRF and tree-structure,
\beq \label{eq:one}
\frac{q^*_v(1)}{q^*_v(0)}& = & \lambda_v
\prod_{\ell=1}^d\frac{\max_{\sigma \in \Sg} \psi_{v_\ell,
v}(\sigma,1)q^*_{\ell}(\sigma)}{\max_{\sigma \in \Sg}
\psi_{v_\ell,v} (\sigma,0)q^*_{\ell}(\sigma)}. \eeq
Now to calculate
$\frac{p^*_v(1)}{p^*_v(0)}$, we define a new graph $G'$ and the
corresponding pair-wise MRF $X'$ as follows. Let $G'$ be the same as
$G$ except that $v$ is replaced by $d$ vertices
$v'_1,v'_2,\ldots,v'_d$; each $v'_\ell$ is connected only to
$v_\ell$, $1\leq \ell \leq d$. The  $X'$ is defined same as $X$
except that $\phi_{v'_\ell}(1)=\lambda_v^{1/d}\phi_v(1)$,
$\phi_{v'_\ell}(0)=\phi_v(0)$ and $\psi_{v_\ell
v'_\ell}=\psi_{v_\ell v}$. Then,
\beq
\frac{p^*_v(1)}{p^*_v(0)} & = &
\frac{\max_{\left\{X':X'_{v'_1}=1,X'_{v'_2}=1,\ldots,X'_{v'_d}=1\right\}}
\Pr_{G'}(X')}{\max_{\left\{X':X'_{v'_1}=0,X'_{v'_2}=0,\ldots,X'_{v'_d}=0\right\}}
\Pr_{G'}(X')} \nonumber \\
 & = & \prod_{\ell=1}^d \frac{\mu_\ell(1)}{\mu_\ell(0)}, \label{eq:2}
 \eeq
 where define
$\mu_\ell(\sigma)=\max_{\{X':X'_{v'_\ell}=\sigma\}} \Pr[X'  |\
X'_{v'_1}=0,\ldots,X'_{v'_{(\ell-1)}}=0,
X'_{v'_{(\ell+1)}}=1,\ldots,X'_{v'_d}=1]$. The second equality in
(\ref{eq:2}) follows by standard trick of Telescoping multiplication
and Lemma \ref{lem:small}.

Now for $1\leq \ell \leq d$, consider MRF $X'(\ell)$ induced on
$G'(\ell) = G'-\{v'_\ell\}$ by fixing \ $\{v'_1,\ldots
v'_d\}-\{v'_\ell\}$ as follows: let $(\phi_{v'_1}(0)=1,
\phi_{v'_1}(1)=0); \ldots; (\phi_{v'_{\ell-1}}(0)=1,
\phi_{v'_{\ell-1}}(1)=0); (\phi_{v'_{\ell+1}}(0)=0,
\phi_{v'_{\ell+1}}(1)=1); \ldots; (\phi_{v'_d}(0)=0,
\phi_{v'_d}(1)=1)$. Then let $\nu_\ell(\sigma), \sigma \in \Sg$
denote the max-marginal of $v_\ell$ for taking value $\sigma$ with
respect to $X'(\ell)$. Given this, by definition of MRF $X'$ as well
$X'(\ell)$ and noting that $v'_\ell$ is a leaf (only connected to
$v_\ell$) with respect to graph $G'$, we have \beq
\frac{\mu_\ell(1)}{\mu_\ell(0)} & = &  \lambda_v^{1/d}
\frac{\max_{\sigma \in \Sg} \psi_{v_\ell, v'_\ell}(\sigma,1)
\nu_\ell(\sigma)} {\max_{\sigma \in \Sg} \psi_{v_\ell,
v'_\ell}(\sigma,0)\nu_\ell(\sigma)}. \label{eq:3} \eeq From
(\ref{eq:one}), (\ref{eq:2}) and (\ref{eq:3}) it is sufficient to
show that \beq \frac{\nu_\ell(1)}{\nu_\ell(0)} & =
&\frac{q^*_\ell(1)}{q^*_\ell(0)}, ~~1\leq \ell \leq d. \label{eq:4}
\eeq Now, note that $T_\ell$ is the same as $\Ts(G(\ell))$ with
respect to $X'(\ell)$. Because for each $\ell=1,\ldots d$,
$G'(\ell)$ has one less \un~node than $G$, the desired result
(\ref{eq:4}) follows by induction hypothesis.
\end{proof}

\begin{lemma}\label{lem:small}
Consider a distribution on $X = (X_1,\dots, X_n)$ where $X_i$ are
binary variables. Let $p_s = \Pr[X = s], s \in \Sg^n$. Let
$p_{s|a_2,\dots,a_d} = \Pr[X=s|X_2=a_2,\dots, X_d = a_d]$ for any $d
\geq 1$. Let $S(a_1,\dots,a_d) = \{ s = (s_1,\dots,s_n) \in \Sg^n :
s_1 = a_1, \dots, s_d = a_d\}$. Then,
$$ \frac{\max_{s \in S(a_1,a_2\dots,a_d)} p_s }{\max_{s \in S(\hat{a}_1,a_2,\dots,a_d)} p_s }  = \frac{\max_{s \in S(a_1,a_2\dots,a_d)} p_{s|a_2,\dots,a_d} }{\max_{s \in S(\hat{a}_1,a_2,\dots,a_d)} p_{s|a_2,\dots,a_d}  }.$$
\end{lemma}
\vspace{.1in}
\begin{proof}
Let $q = \Pr(X_2=a_2,\dots,X_d=a_d)$. Then, by definition of
conditional probability for $s \in S(a_1, a_2,\dots, a_d) \cup
S(\hat{a}_1,a_2,\dots,a_d)$, $ p_s = p_{s|a_2,\dots,a_d} q.$ From
this, Lemma follows immediately.
\end{proof}

\subsection{Size of Self-avoiding walk tree}

We present a novel characterization of the size of the self avoiding walk tree in
terms of number of edges in it (which is equal to number of nodes minus 1).
This characterization is necessary to obtain bound on the running time of
the self-avoiding walk tree. This combinatorial result should be of interest
in its own right.

\begin{lemma}\label{lem:tsaw}
Consider a connected graph $G=(V,E)$ with $|V|=n$ nodes and
$|E|=n-1+k$ edges, $k \geq 0$. Then for any $v \in V$,
$ |\Ts(G,v)| \leq (n+k-1) 2^{k+1}.$
Further, there exists a graph with $n-1+k$ edges with $k < n/2$
so that for any node $v \in V$,
$ |\Ts(G,v)| \geq n 2^{k-2}.$
\end{lemma}
\begin{proof}
The proof is divided into two parts. We first provide the proof of
lower bound. Consider a line graph of $n$ nodes (with $n-1$ edges).
Now add $k < n/2$ edges as follows. Add an edge between $1$ and $n$.
Remaining $k-1$ edges are added between node pairs:
$(2,4),(4,6),\dots,(2(k-2),2(k-1)),(2(k-1), 2k)$. Consider any node,
say $v$.  It is easy to see that there are at least $2^{k-2}$
different ways in which one can start walking on the graph from node
$v$ towards node $1$, cross from $1$ to $n$ via edge $(1,n)$ and
then come back to node $v$. Each of these different loops, starting
from $v$ and ending at $v$ creates $2$ distinct paths in the
self-avoiding walk tree of length at least$\frac{n}{2}$. Thus, the
size of self-avoiding walk tree of each node is at least $n 2^{k-2}$
for each node.  This completes the proof of lower bound.

Now, we prove the upper bound of $n 2^{k+1}$ on the size of
self-avoiding walk tree for each node $v \in V$. Given that $G$ is
connected, we can divide the edge set $E = E_T \cup  E_k$ where $E_k
= \{e_1,\dots,e_k\}$ and $T = (V, E_T)$ forms a spanning tree of
$G$. Let $\cS$ be the set of all subsets of $E_k =
\{e_1,\dots,e_k\}$ (there are $2^k$ of them including empty set).
Now fix a vertex $v \in V$ and we will concentrate on $\Ts(G,v)$.
Consider any $u \in V$ (can be $v$) and  $S \in \cS$. Next, we wish
to count number of paths in $\Ts(G,v)$ that end at (a copy of) $u$
(however, $u$ need not be a leaf), contain all edges in $S$ but none
from $E_k \backslash S$. We claim the following.

{\em Claim. } There can be at  most one path of $\Ts(G,v)$ from $v$
to (a copy of) $u$ and containing all edges from $S$ but none from
$E_k \backslash S$.

\begin{proof}
To prove the above claim, suppose it is not true. Then there are at
least two distinct paths from $v$ to $u$ that contain all edges in
$S$ (but none from $E_k \backslash S$). Consider the symmetric
difference of these two paths (in terms of edges). This symmetric
difference must be a non-empty subset of $E_T$ and also contain a
loop (as the two paths have same starting and ending point). But
this is not possible as $T=(V, E_T)$  is a tree and it does not
contain a loop. This contradicts our assumption and proves the
claim.
\end{proof}

Given the above claim, for any node $u$, clearly the number of
distinct paths from node $v$ to (a copy of) $u$ in $\Ts(G,v)$ are at
most $2^k$. Now each edge has two end points. For each appearance of
an edge of $G$ in $\Ts(G,v)$, a distinct path from $v$ to one of its
end point must appear in $\Ts(G,v)$. From above claim, this can
happen at most $2 \times 2^k = 2^{k+1}$. There are $n+k-1$ edges of
$G$ in total. Thus, net number of edges that can appear in
$\Ts(G,v)$ is at most $(n+k-1)2^{k+1}$; thus completing the proof of
Lemma \ref{lem:tsaw}.
\end{proof}

\subsection{Algorithm: At a higher level}

Now, we describe algorithm to compute MAP approximately. The
algorithm is the same as \MAP, ~however computation restricted
to each component is done through self-avoiding walk. Specifically,
the algorithm does the following: given $G$,
decompose it into (small) components $S_1,\dots,S_K$ by removing
(few) edges $\cB \subset E$, where $\cB$ is obtained using \DEC;
(as before, for minor-excluded graph use \DCe ~ and \DD ~ for graphs
with low doubling dimension).  Then, compute an approximate MAP
assignment by computing exact MAP restricted to the components. This
exact computation for each component is performed through a message
passing mechanism using the equivalence stated in Theorem \ref{thm:two}:
essentially, growing self-avoiding walk tree is just sending messages
along a breadth-first search tree; computation over a self-avoiding walk
tree is essentially standard max-product (message passing) algorithm.
The precise schedule for message-passing is described in the next
sub-section. Here, we describe algorithm for any graph $G$ at a
higher-level.

\vspace{.02in} \noindent \MAP($G$) \ \vspace{.05in} \hrule
\vspace{.05in}
\begin{itemize}
\item[(1)] Use \DEC($G$) to obtain $\cB \subset E$ such that
\begin{itemize}
\item[(a)] $G' = (V, E \backslash \cB)$ is made of connected
components $S_1,\dots, S_K$.
\end{itemize}

\item[(2)] For each connected component $S_j, 1\leq j\leq K$, do the
following:
\begin{itemize}
\item[(a)] Compute exact MAP $\bx^{*,j}$ for component $S_j$, where
$\bx^{*,j}=(x^{*,j}_i)_{i\in S_j}$.
\item[(b)] Computation of $\bx^{*,j}_i$ is performed by growing self-avoiding
walk tree at node $i$  restricted to induced graph by nodes
of $S_j$ using a message passing mechanism; then computing
max-marginal on self-avoiding walk tree using message passing
mechanism (i.e. standard max-product algorithm on self-avoiding
walk tree).

\end{itemize}

\item[(3)] Produce output $\widehat{\bx^*}$, which is obtained by
assigning values to nodes using $\bx^{*,j}, 1\leq j\leq K$. This is
clearly local operation.

\end{itemize}

\vspace{.02in} \hrule \vspace{.05in}

\subsection{Algorithm: Message-passing  schedule}

The following is a pseudo-code of a distributed message passing
algorithm \M ~ which computes $\bx^{*,j}$ for each component $S_j$. The \M ~ finds exact MAP, by Theorem \ref{thm:two}.
This section is of interest primarily for the reason that it provides
the detailed distributed message-passing implementation for computing
MAP. A reader, not interested in such detailed implementation, may
skip this section.

To describe the pseudo-code,  we need some notation. Each
node $v \in V$, let $N(v)$ denote the set of all its neighbors,
i.e. $N(v) = \{ u \in V: (u, v) \in E\}$. Node $v$  assigns
an arbitrary fixed order to all nodes in $N(v)$. For example,
if $v$ has neighbors $u, w$ and $z$ then it can number
$u$ as the first neighbor, $w$ as second neighbor
and $z$ as third neighbor. The ordering chosen by each
node is independent of choices of all other nodes. The
algorithm operates in two phases.  In the first phase,
algorithm explores  local topology for each node
via sending ``path sequences''. By ``path sequence'' we mean a
finite sequence of vertices $(v_1,v_2,\ldots, v_k),$ where
$(v_\ell, v_{\ell+1}) \in E$ for $1\leq \ell \leq k-1$.
In the second phase, algorithm uses the path sequences
to recursively calculate ``computation sequence'' which
in turn leads to calculation of $q^*_v(\cdot)$ at nodes.
A ``computation sequence'' is of the form $(v_1,v_2,\ldots,
v_k, m_{v_k}(0),m_{v_k}(1)),$ where $m_{v_k}(\cdot)$ are
certain real-numbers (which have interpretation of message).
As we shall see, the structure of recursive calculation to
obtain ``computation sequence'' is the same as that of max-product
algorithm. Thus, there is very strong connection between MP
and \M. For ease of exposition, the algorithm
is described to compute the ratio $q^*_v(1)/q^*_v(0)$
for all $v \in V$.

\vspace{.05in} \noindent{\M(G) }\ \vspace{.05in} \hrule
\vspace{.1in}
\begin{itemize}

\item[(0)] Initially, each vertex $v$ sends a path sequence $(v)$ to
each of its neighbors.

\item[(1)] When node $u$ receives a path sequence $(v_1,v_2,\ldots,
v_k)$ from its neighbor $v$, (note that, by construction given
later, $v_{k}=v$) it does the following:
\begin{itemize}
\item[$\circ$] If $u$ is a leaf (i.e. $u$ is connected only to $v$), $u$ sends
back a computation sequence $(v_1,v_2,\ldots, v_k,u,m_u(0),m_u(1))$ to
$v$, where
\beq
m_u(\sigma) & \propto & \max_{\sigma_u \in \Sg} \psi_{u,v}(\sigma_u, \sigma) \phi_u(\sigma_u) \nonumber \\
\sum_{\sigma_u \in \Sg} m_u(\sigma_u) & = & 1.
\eeq

\item[$\circ$] If $u$ is not a leaf, check whether
$u$ appears among $v_\ell$, $1\leq \ell \leq  k$:
\begin{itemize}
\item 
If NO, $u$ sends a path sequence $(v_1,\ldots,v_k,u)$
to each of $u$'s neighbors but $v$.

\item If YES, then let $v_\ell = u, 1\leq \ell < k$.
\begin{itemize}
\item[$-$] If, with respect to the ordering given by
node $u$ to its neighbors, the rank (order) of  node $v_{\ell+1}$ is
larger then $v$, then $u$ sends back (to $v$) a computation sequence
$(v_1,v_2,\ldots, v_k,u,m_u(0),m_u(1))$, where $m_u(1) = 1$ and
$m_u(0) = 0$.

\item[$-$] Otherwise (i.e. the rank of node $v_{\ell+1}$ is smaller than $v$),
$u$ sends back (to $v$) computation sequence $(v_1,v_2,\ldots,
v_k,u,m_u(0),m_u(1))$, where $m_u(0) = 1$ and $m_u(1) = 0$.

\end{itemize}

\end{itemize}

\end{itemize}

\item[(2)] Once a node $u$ receives a computation sequence
$(v_1,\ldots,v_k,m_{v_k}(0), m_{v_k}(1))$ from its neighbor $v$,
(note that, by construction $v_{k}=v$ and $v_{k-1}=u$).
Store this computation sequence in $u$'s memory and do
the following:

\begin{itemize}
\item[$\circ$] If $k>2$, check whether $u$ has stored computation sequences of
the form $(v_1,\ldots,v_{k-1},w,m_w(0),m_w(1))$ for all
$w \in N(u)-\{v_{k-2}\}$. If so, $u$ sends a computation sequence
$(v_1,\ldots,v_{k-1} (=u),m_u(0),m_u(1))$ to $v_{k-2}$
where
\beq
m_u(\sigma) & \propto & \left[ \max_{\sigma_u \in \Sg} \psi_{u,v_{k-2}}(\sigma_u, \sigma) \phi_u(\sigma_u) \times  \rf. \nonumber \\
& ~ & ~~~~\lf. \prod_{w \in N(u) - \{ v_{k-2} \}} m_{w}(\sigma_u)\right], \nonumber
\eeq
\beq
\sum_{\sigma_u \in \Sg} m_{u}(\sigma_u) & = & 1. \nonumber
\eeq
Delete computation sequences $(v_1,\ldots,v_{k-1},w,m_w(0),m_w(1))$
for all $w\in N(j)-\{i_{k-2}\}$ from $u$'s memory.

\item[$\circ$] If $k=2$, then check whether for all $w \in N(j)$, $u$
has stored computation sequences $(v_1,w,m_w(0),m_w(1))$.
If so, compute the (estimate of) max-belief of $u$ as
$$q^*_u(\sigma) \propto \phi_u(\sigma) \prod_{w \in N(u)} m_{w}(\sigma),
\mbox{~~~and~~~} \sum_{\sigma \in \Sg} q^*_u(\sigma) = 1.$$
\end{itemize}

\item[(3)] When all nodes have computed their max-beliefs,
declare $q^*_v(1)/q^*_v(0)$ as an estimate of $p^*_v(1)/p^*_v(0)$
$ \forall ~v \in V$.

\end{itemize}
\vspace{.1in} \hrule \vspace{.1in}

\section{Experiments}\label{sec:exp}

Our algorithm provides provably good approximation
for any MRF that has low doubling dimension or that excluded
minor. The planar graph is a special case of such graphs. The popular
model of grid graph, which is both planar and has low doubling dimension,
will be used in the experimental section. We will, however, use
the decomposition algorithm \DCe~for obtaining our results.
Now we present detailed setup and experimental results.

\subsection{Setup 1}

Consider\footnote{Though this setup has $\phi_i,
\psi_{ij}$ taking negative values, they are equivalent to the setup
considered in the paper as the function values are lower bounded and
hence {\em affine} shift will make them non-negative without
changing the distribution.}
 binary (i.e. $\Sigma =
\{0,1\}$) MRF on an $n \times n$ lattice $G = (V,E)$: {\small $$
\Pr(\bx) \propto \exp\lf(\sum_{i\in V} \theta_i x_i + \sum_{(i,j)
\in  E} \theta_{ij} x_i x_j\rf), ~~\mbox{for} ~\bx \in
\{0,1\}^{n^2}. $$}

\begin{figure}[htb]
\begin{center}
\begin{psfrags}
\includegraphics[width=5cm]{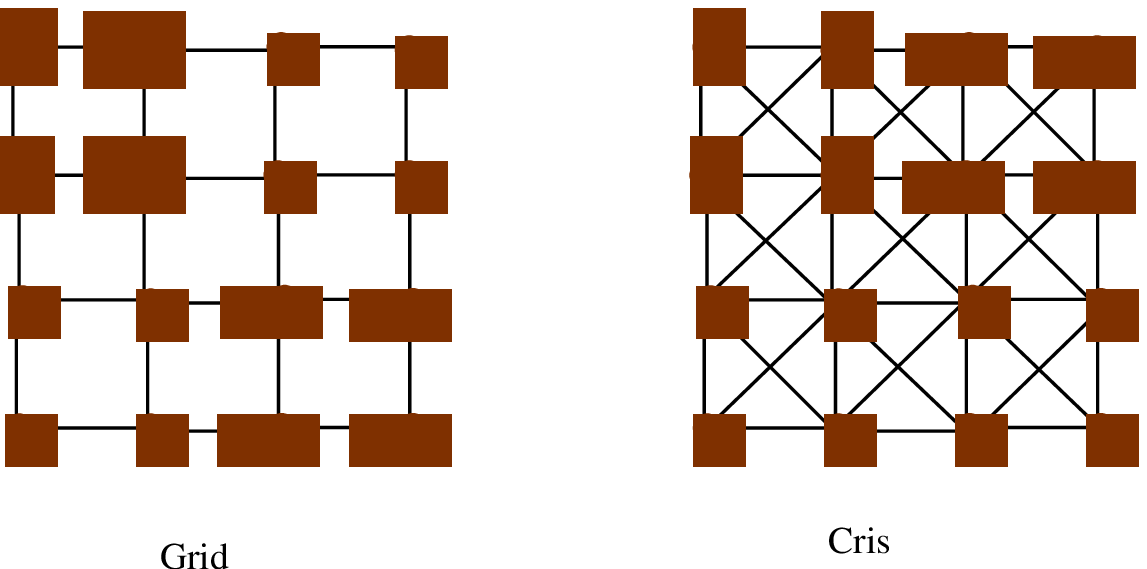}
\end{psfrags}
\end{center}
\caption{Example of grid graph (left) and cris-cross graph (right)
with $n=4$.} \label{fig:one} \vspace{-.1in}
\end{figure}

Figure \ref{fig:one} shows a lattice or grid
graph with $n = 4$ (on the left side). There are two scenarios for
choosing parameters (with notation $\cU[a,b]$ being uniform
distribution over interval $[a,b]$):

(1) {\em Varying interaction.} $\theta_i$ is chosen independently
from distribution $\cU[-0.05,0.05]$ and $\theta_{ij}$ chosen
independent from $\cU[-\alpha, \alpha]$ with $\alpha \in
\{0.2,0.4,\dots, 2\}$.

(2) {\em Varying field.} $\theta_{ij}$ is chosen independently from
distribution $\cU[-0.5,0.5]$ and $\theta_{i}$ chosen independently from
$\cU[-\alpha, \alpha]$ with $\alpha \in \{0.2,0.4,\dots, 2\}$.

\begin{figure}[htb]
\begin{center}
\begin{psfrags}
\includegraphics[width=8.5cm,height=8cm]{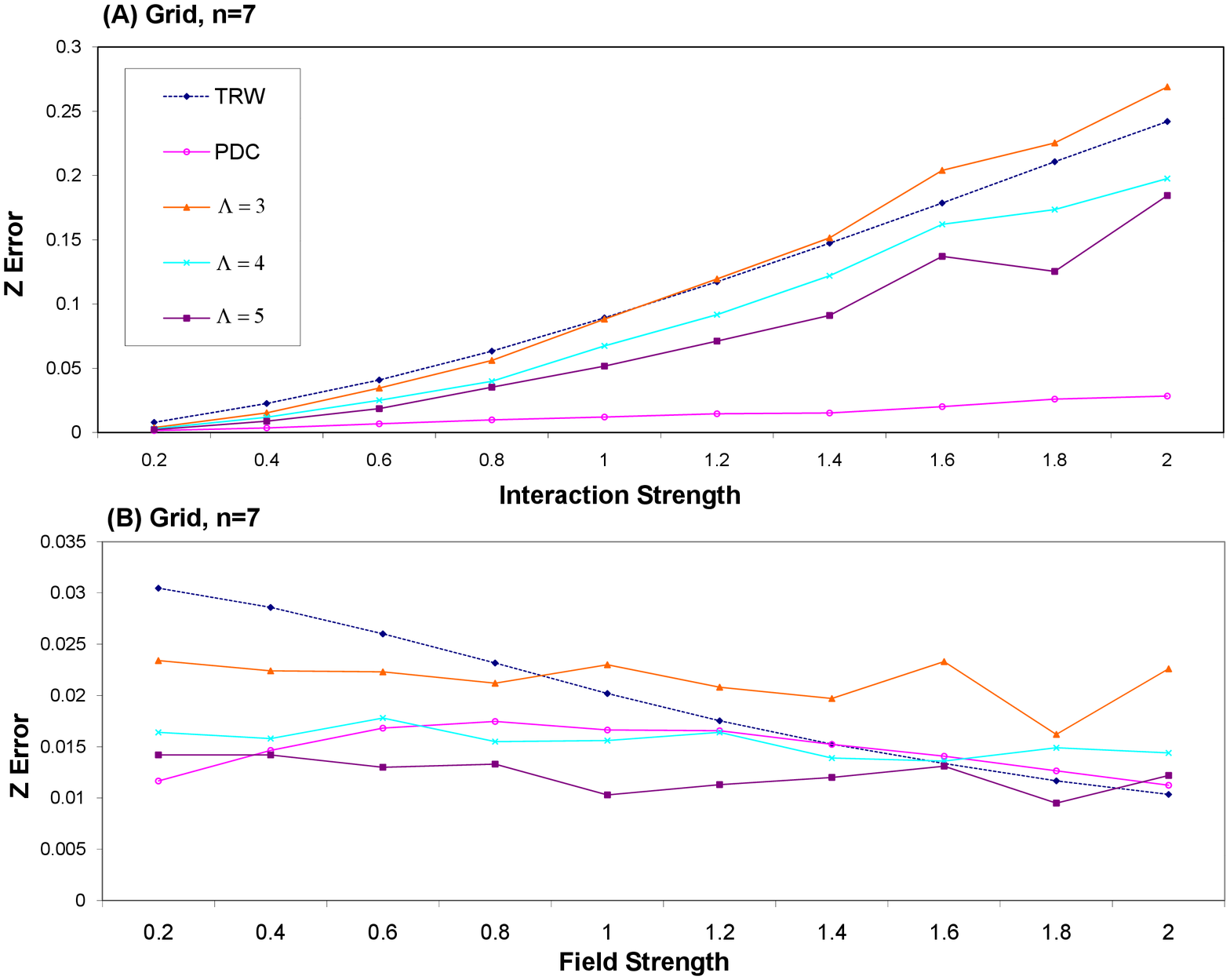}
\end{psfrags}
\end{center}
\caption{\small Comparison of TRW, PDC and our algorithm for grid
graph with $n=7$ with respect to error in $\log Z$. Our algorithm
outperforms TRW and is competitive with respect to PDC. }
\label{fig:r1} \vspace{-.1in}
\end{figure}

The grid graph is planar. Hence, we run our algorithms \LOGP~ and
\MAP, with decomposition scheme \DC($G, 3, \bDelta$),
$\bDelta \in \{3, 4, 5\}$. We consider two measures to evaluate
performance: error in $\log Z$, defined as
$\frac{1}{n^2} | \log Z^{\text{alg}} -\log Z |$; and
error in $\E(\bx^*)$, defined as $\frac{1}{n^2} |\E(\bx^{\text{alg}} - \E(\bx^*)|$.

We compare our algorithm for error in $\log Z$ with the two recently
very successful algorithms -- Tree re-weighted algorithm (TRW) and
planar decomposition algorithm (PDC). The comparison is plotted in
Figure \ref{fig:r1} where $n = 7$ and results are averages over $40$
trials. The Figure (A) plots
error with respect to varying interaction while   Figure (B) plots
error with respect to varying field strength. Our algorithm, essentially
outperforms TRW for these values of $\bDelta$ and perform very competitively
with respect to PDC.

\begin{figure}[htb]
\begin{psfrags}
\includegraphics[width=8.5cm,height=10cm]{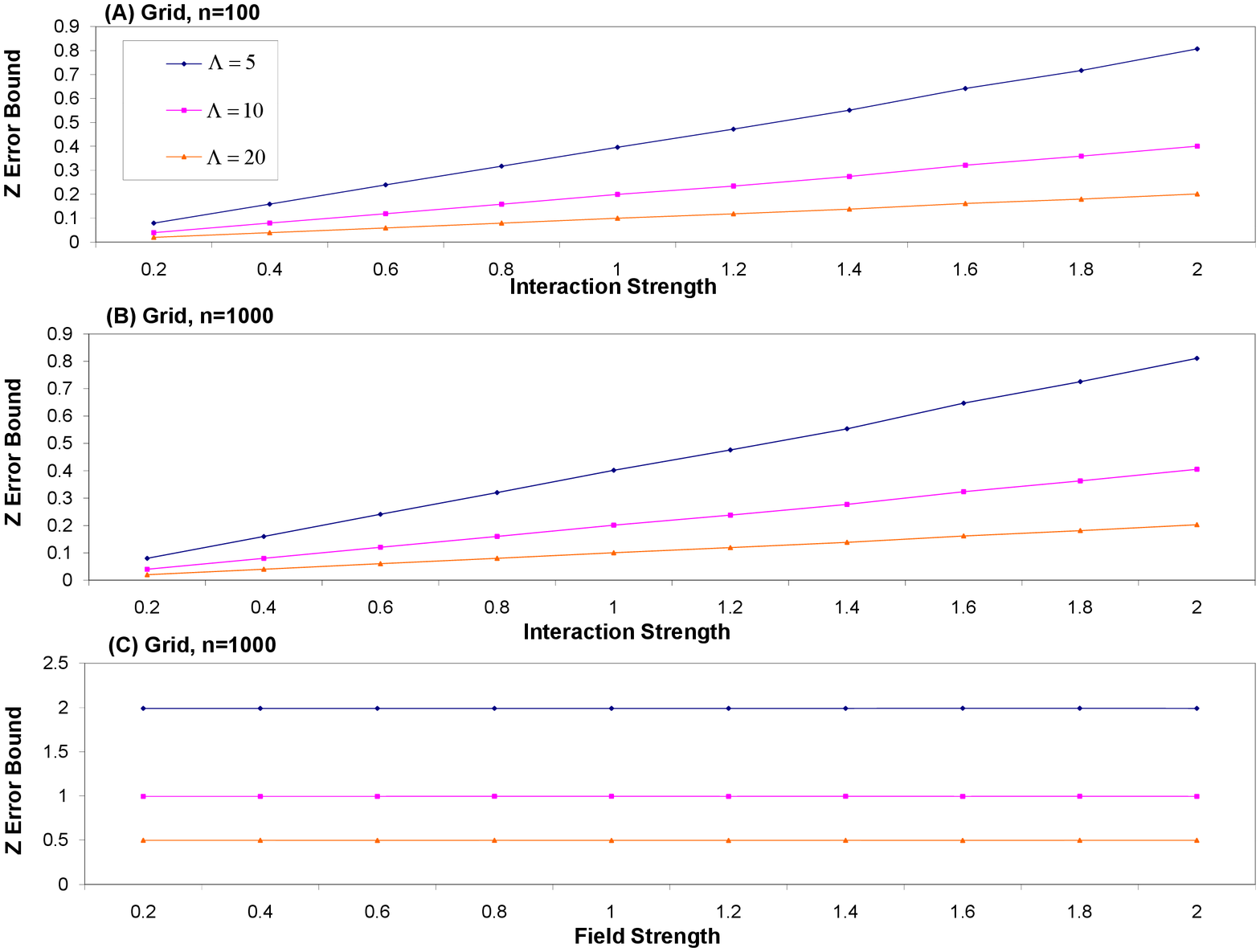}
\end{psfrags}
\caption{\small The theoretically computable error bounds for $\log
Z$ under our algorithm for grid with $n=100$ and $n=1000$
under varying interaction and varying field model. This clearly
shows scalability of our algorithm.} \label{fig:r2} \vspace{-.1in}
\end{figure}

The key feature of our algorithm is scalability. Specifically,
running time of our algorithm with a given parameter value $\bDelta$
scales linearly in $n$, while keeping the relative error bound
exactly the same. To explain this important feature, we plot the
theoretically evaluated bound on error in $\log Z$ in Figure
\ref{fig:r2} with tags (A), (B) and (C). Note that error bound plot
is the same for $n=100$ (A) and $n=1000$ (B). Clearly, actual error
is likely to be smaller than these theoretically plotted bounds. We
note that these bounds only depend on the interaction strengths and
{\em not} on the values of fields strengths (C).

\begin{figure}[htb]
\begin{psfrags}
\includegraphics[width=8.5cm,height=10cm]{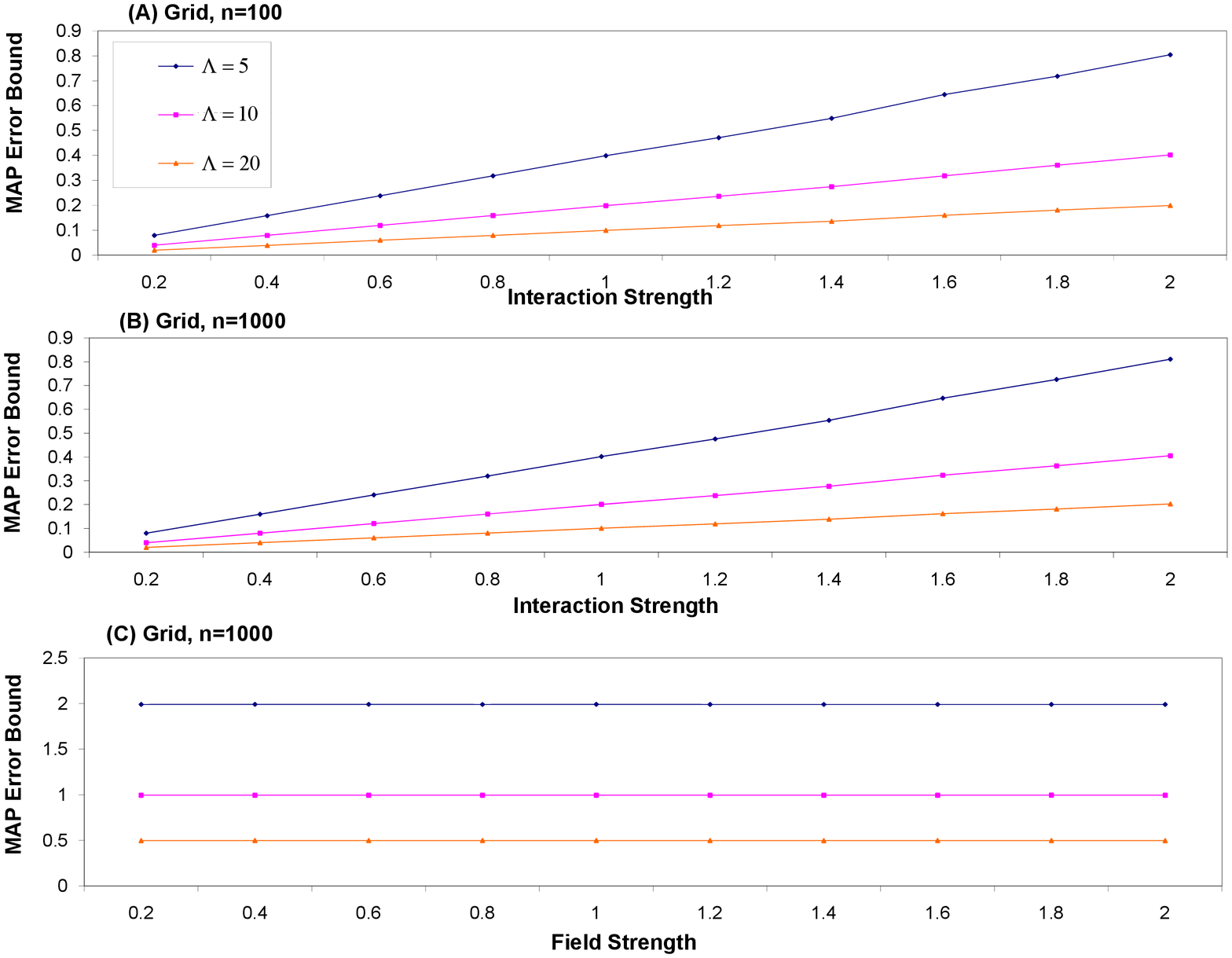}
\end{psfrags}
\caption{\small The theoretically computable error bounds for MAP under our algorithm for grid with $n=100$ and $n=1000$
under varying interaction and varying field model.} \label{fig:r3} \vspace{-.1in}
\end{figure}

Results similar to of \LOGP~are expected from \MAP. We plot the
theoretically evaluated bounds on the error in MAP in Figure
\ref{fig:r3} with tags (A), (B) and (C). Again, the bound on MAP
relative error for given $\bDelta$ parameter remains the same for all
values of $n$ as shown in (A) for $n=100$ and (B) for $n = 1000$.
There is no change in error bound with respect to the field strength
(C).

\subsection{Setup 2}

\begin{figure}[htb]
\begin{center}
\begin{psfrags}
\includegraphics[width=8.5cm,height=10cm]{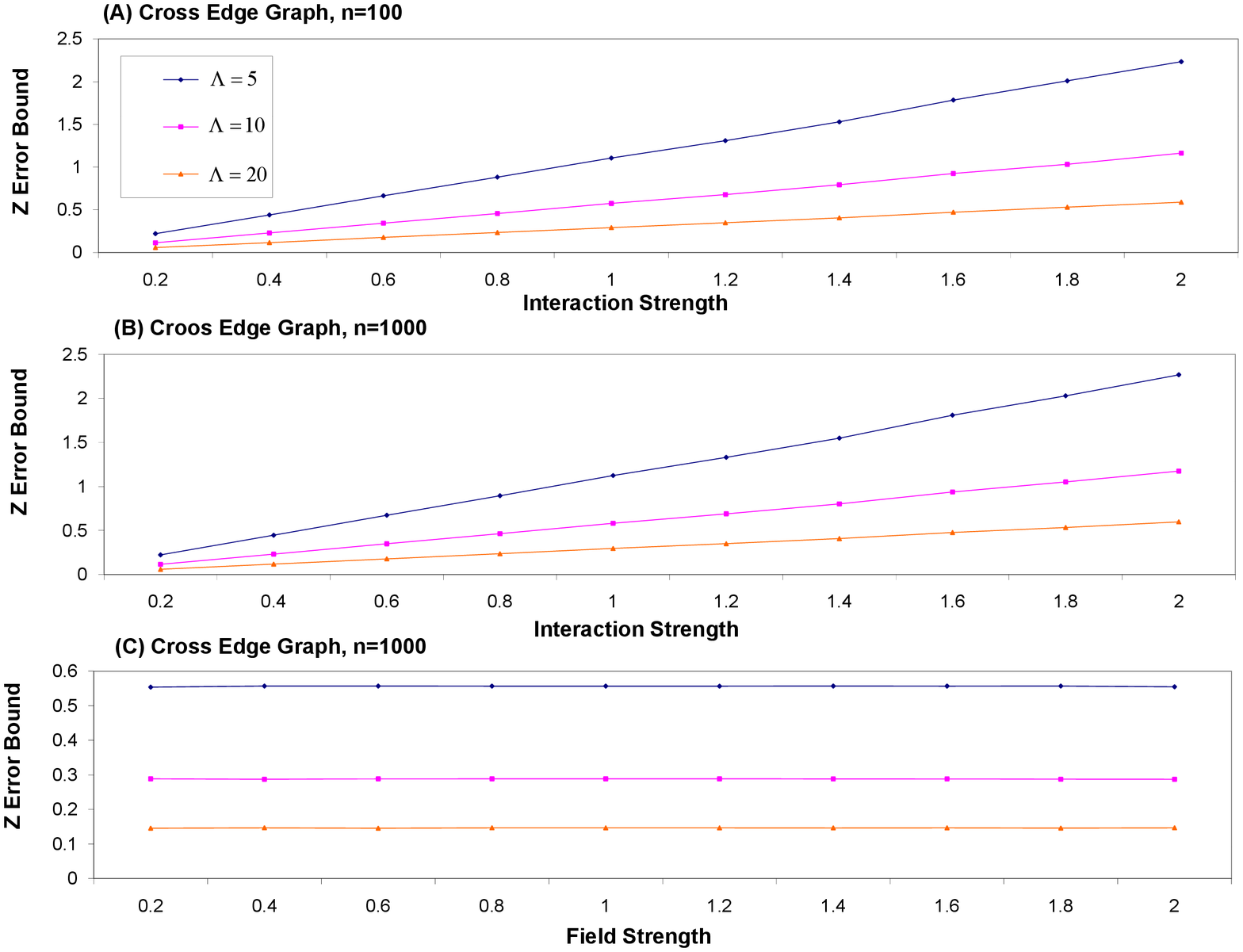}
\end{psfrags}
\end{center}
\caption{\small The theoretically computable error bounds for $\log
Z$ under our algorithm for cris-cross with $n=100$ and
$n=1000$ under varying interaction and varying field model. This
clearly shows scalability of our algorithm and robustness to graph
structure.} \label{fig:r4} \vspace{-.1in}
\end{figure}

\begin{figure}[htb]
\begin{center}
\begin{psfrags}
\includegraphics[width=8.5cm,height=10cm]{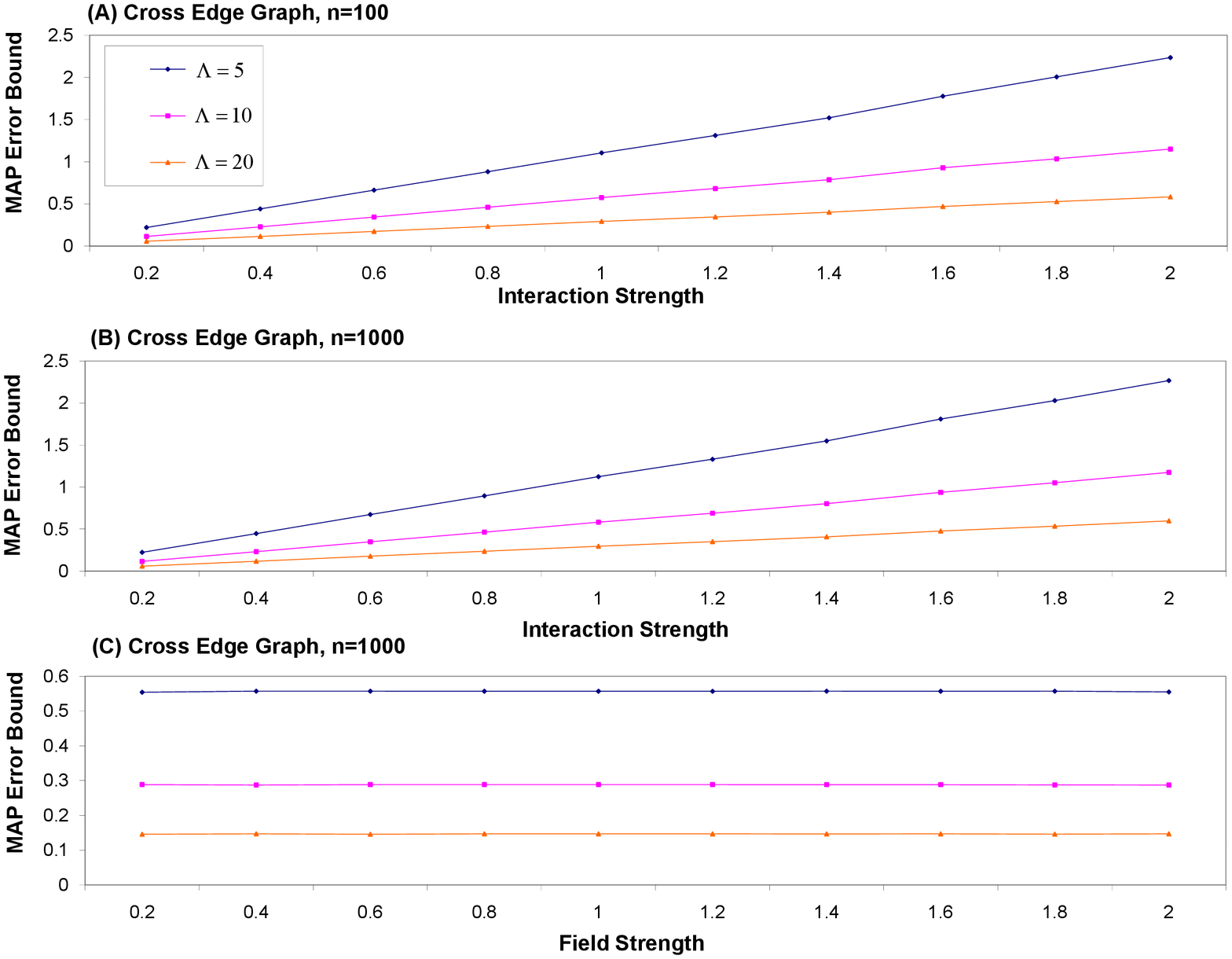}
\end{psfrags}
\end{center}
\caption{\small The theoretically computable error bounds for MAP under our algorithm for cris-cross with $n=100$ and
$n=1000$ under varying interaction and varying field model.} \label{fig:r5} \vspace{-.1in}
\end{figure}

Everything is exactly the same as the above setup
with the only difference that grid graph is replaced by {\em
cris-cross} graph which is obtained by adding extra four neighboring
edges per node (exception of boundary nodes). Figure \ref{fig:one}
shows cris-cross graph with $n=4$ (on the right side). We again run
the same algorithm as above setup on this graph. For cris-cross
graph, which is graph with low-doubling dimension, we obtained
its graph decomposition from the decomposition of its grid sub-graph. Therefore
, the running time of our algorithm remains the same
(in order) as that of grid graph and error bound will become
only $3$ times weaker than that for the grid graph. We compute
these theoretical error bounds for $\log Z$ and MAP which is
plotted in Figure \ref{fig:r4} and \ref{fig:r5}. These figures are similar to the Figures \ref{fig:r2} and \ref{fig:r3}
for grid graph.

\section{Unexpected implication: existence of limit}\label{sec:imp}

This section describes an important and somewhat unexpected
implication of our results, specifically Lemmas \ref{lem:main1b}
and \ref{lem:main2b}. In the context of {\em regular} MRF, such
as an MRF on $\bbZ_n^d$ (of $n^d$ nodes) with same node and edge potential
functions for all nodes and edges, we will show that (non-trivial)
limit $\frac{1}{n^d} \log Z$ exists as $n\to \infty$. It is worth noting that
showing existence of such limits is not straightforward in general
and hence our method should be of interest as such an analytic
tool. We believe that the result stated below is well-known; however
its proof method is likely to allow for establishing such existence
for a more general class of problems. As an example, the theorem
will hold even when node and edge potentials are not the same
but are chosen from a class of such potential as per some distribution
in an i.i.d. fashion.  Now, we state the result.

\begin{theorem}\label{thm:limit}
Consider a regular MRF of $n^d$ nodes on $d$-dimensional
grid $\bbZ_d^n = (V_n, E_n)$: let $\psi_{ij} \equiv \psi, \phi_i \equiv \phi$ for
all $i \in V_n, (i,j) \in E_n$ with $\psi : \Sigma^2 \to \Rp, \phi: \Sigma \to \Rp$.
Let $Z_n$ be partition function of this MRF. Then, the following
limit exists:
$$ \lim_{n\to\infty} \frac{1}{n^d} \log Z_n = A(d, \phi,\psi) \in (0,\infty).$$
\end{theorem}

\subsection{Proof of Theorem \ref{thm:limit}}

The proof of Theorem \ref{thm:limit} is stated for $d = 2$ and $\Sigma = \{0,1\}$
case. Proof for $ d \geq 3$ and $\Sigma$ with $\Sgl \geq 2$ can be proved using
exactly the same argument. The proof will use the following Lemmas.

\begin{lemma}\label{lem:lim1}
Let $d=2$ and $\phi^* = \max_{\sigma \in \{0,1\}} \phi(\sigma)$,
$\psi^* = \max_{(\sigma, \sigma') \in \{0,1\}^2} \psi(\sigma,\sigma').$ Then,
$$ n^2 \leq \log Z_n \leq \alpha n^2, $$
where $\alpha = \log 2 + \log \phi^* + 4 \log \psi^*$.
\end{lemma}

\begin{lemma}\label{lem:lim2}
Define $a_n = \frac{1}{n^2} \log Z_n$.  Now, given $k > 0$, there exists $n(k)$
large enough such that for any $m, n \geq n(k)$,
$$ | a_m - a_n | = O\lf(\frac{1}{k}\rf) + O\lf(\frac{k}{\min{\{{m},{n}}\}}\rf).$$
\end{lemma}

\begin{proof}{\em (Theorem \ref{thm:limit})} We state proof of Theorem \ref{thm:limit},
before proving the above stated Lemmas. First note that, by Lemma \ref{lem:lim1},
the elements of sequence $a_n = n^{-2} \log Z_n$ take value in $[1,\alpha]$.
Now, suppose the claim of theorem is false. That is, sequence $a_n$ does not
converge as $n\to \infty$. That is, there exists $\delta > 0$ such for any
choice of $n_0$, there are $m \geq n \geq n_0$ such that
$$ |a_m - a_n | \geq \delta.$$
By Lemma \ref{lem:lim2}, we can select $k$ large enough and later $n_0 \geq n(k)$ large
enough such that for any $m, n \geq n_0$,
$$ |a_m - a_n | < \delta.$$
But this is a contradiction to our assumption that $a_n$ does not converge to a limit.
That is, we have established that $a_n$ converges to a non-trivial limit in $[1,\alpha]$
as desired. This completes the proof of Theorem \ref{thm:limit}.
\end{proof}

\subsection{Proofs of Lemmas}

\begin{proof}{\em (Lemma \ref{lem:lim1})}
Consider the following.
\beq
2^{n^2} & = & \sum_{\bx \in \{0,1\}^{n^2}} \prod_{i \in V_n} 1 \prod_{(i,j) \in E_n} 1 \nonumber \\
               & \stackrel{(a)}{\leq} & \sum_{\bx \in \{0,1\}^{n^2}} \prod_{i \in V_n} \exp(\phi(x_i)) \prod_{(i,j) \in E_n} \exp(\psi(x_i,x_j)) \nonumber \\
               & = & Z_n \nonumber \\
               & \stackrel{(b)}{\leq} & \sum_{\bx \in \{0,1\}^{n^2}} \prod_{i \in V_n} \exp(\phi^*) \prod_{(i,j) \in E_n} \exp(\psi^*).
\eeq
Here, (a) follows from the fact that $\psi, \phi$ are non-negative valued functions and
(b) follows from definitions of $\phi^*, \psi^*$.  Now, taking logarithm on both sides implies
the Lemma \ref{lem:lim1}.
\end{proof}

\begin{proof}{\em (Lemma \ref{lem:lim2})}
Given $k > 0$, consider $n$ large enough (will be decided later). Consider $\bbZ_n^2 = (V_n, E_n)$
and let it be laid out on $X-Y$ plane so that its node in $V_n$ occupy the integral
locations : $(i,j), ~0\leq i\leq n-1, 0\leq j\leq n-1$. Now, we describe a scheme to
obtain a $(O(1/k), O(k^2))$ edge-decomposition of $\bbZ^2_n$. For this,
choose $\ell_1, \ell_2 \in \{0,\dots, k-1\}$ independently and uniformly at
random. Select edges to form $\cB$ to obtain edge-decomposition as
follows: select vertical edges with bottom vertex having $Y$ coordinate
$\ell_2 + j k, j \geq 0$, and select horizontal edges with left vertex having
$X$ coordinate $\ell_1 + j k, j \geq 0$. That is,
\beq
 \cB & = & \{(u,v) \in E_n : u = (i, j), v=(i+1, j), i \mbox{~mod~} k = \ell_1 \}~~~~ \nonumber \\
& \cup & \{(u,v) \in E_n : u = (i, j), v=(i, j+1), j \mbox{~mod~} k = \ell_2 \}.~~~~\nonumber
\eeq
It is easy to check that this is $(O(1/k), O(k^2))$ edge-decomposition due to uniform
selection of $\ell_1, \ell_2$ from $\{0,\dots,k-1\}$. Therefore, by Lemma \ref{lem:main1b},
we can obtain estimates that are $(1\pm O(1/k)) \log Z_n$ using our algorithm.

Let $m = \lceil n/k \rceil$.  Under the  decomposition $\cB$ as described above, there are
at least $(m-1)^2$ connected components that are MRF on $\bbZ_k$. Also, all the
connected components can be covered by at most $(m+1)^2$ identical MRFs on $\bbZ_k$.
Using arguments similar to those employed in calculations of Theorem \ref{lem:main1} (using
non-negativity of $\phi, \psi$), it can be shown that the estimate produced by our algorithm
is lower bounded as
\beq
(1- O(1/k)) (m-1)^2  \log Z_k & = & n^2 \frac{\log Z_k}{k^2} \times ~~~\nonumber \\
& ~ & \lf( 1 - O(1/k) - O(k/n) \rf), ~~~\nonumber
\eeq
and is upper bounded as
\beq
(1+ O(1/k)) (m+1)^2 \log Z_k & = & n^2 \frac{\log Z_k}{k^2} \times~~~ \nonumber \\
 & ~ & \lf( 1 + O(k/n) + O(1/k) \rf).~~~\nonumber
\eeq
Therefore, from above discussion we obtain that
$$ \frac{1}{n^2} \log Z_n =  \frac{Z_k}{k^2} \lf( 1 \pm O(k/n) \pm O(1/k)\rf).$$
Therefore, recalling notation of $a_n$, we have that
$$ |a_m - a_n | = a_k O\lf(\frac{k}{\min{\{m,n\}}}\rf) + a_k O(1/k).$$
Since, $a_k \in [1,\alpha]$ for all $k$, we obtain the desired result of Lemma \ref{lem:lim2}.
\end{proof}

\section{Conclusion}

In this paper, we present simple novel local approximation algorithm for
computing log-partition function and MAP estimation for arbitrary exponential
distribution represented by a pair-wise MRF. We showed these algorithms
provide bounds for arbitrary graph with quantifiable approximation guarantees.
Further, for low-doubling dimension graphs and minor-excluded graphs
it can provide arbitrary accuracy within linear time. The main takeaway
for a practitioner is the following: there is a simple and intuitive local algorithm
that provides provable bounds with computable approximation error
for any graph and hence it can be used as a good heuristic and producing
approximation guarantee certificate.

We proposed message-passing implementation based on self-avoiding walk
trees which should provide such implementation for other problems as well.
This method, through a transformation from non-binary exponential family to
binary MRF, extends for any finite valued factor graph. However, this can result
in somewhat redundant construction. Understanding  design of direct constructions
for non-binary pair-wise MRF is an important open problem.

We derived an unusual implication of our algorithmic results for providing
existence of asymptotic limits of free energy for a class of regular MRFs. Our
result suggest a way to explicitly evaluate these limiting up to an arbitrary
accuracy. This should be of general interest as a method for establishing
asymptotic limits as well as computing these limits.

Finally, we remark that our methods are explained for exponential family
only. However, they easily extend to certain hard-core models such as
independent set or matching where there is a {\em non-constraining} assignment
to node values.


\bibliographystyle{IEEEtran}
\bibliography{IEEEabrv,inference}

\appendices

\section{Proof of Lemma \ref{lem:poly}}\label{ap:01}

The proof is
by induction on $r \in \N$. For base case, consider $r = 0$. Now,
$\bB(x, 2^0=1)$ is essentially the set of all points which are at
distance $< 1$ by definition. Since it is metric with distance being
integer, this means that the set of all points that are at distance
$0$. By definition of metric, we have that $x$ is the only such
point. That is, $\bB(x,1) = \{x\}$. Hence, $|\bB(x, 1) | = 1 \leq
2^{0 \times \rho(\cM)}$ for all $x \in \cX$.

Now suppose the claim of Lemma is true for all $r \leq k$ and all
$x\in \cX$. Consider $r = k+1$ and any $x \in \cX$. By definition of
doubling dimension, there exists $\ell \leq 2^{\rho(\cM)}$ balls of
radius $2^k$, say $\bB(y_j, 2^k)$ with $y_j \in \cX$ for $1\leq
j\leq \ell$, such that
$$ \bB(x, 2^{k+1}) \subset \cup_{j=1}^\ell \bB(y_j, 2^k).$$
Therefore,
$$ |\bB(x,2^{k+1}) | \leq \sum_{j=1}^\ell |\bB(y_j, 2^{k})|.$$
By inductive hypothesis, for $1\leq j\leq \ell$,
$$|\bB(y_j, 2^k)| \leq  2^{k \rho(\cM)}.$$
Since we have $\ell \leq 2^{\rho(\cM)}$, we obtain
$$ |\bB(x,2^{k+1})| \leq \ell ~ 2^{k\rho(\cM)} ~\leq~ 2^{(k+1)\rho(\cM)}.$$
This completes the proof of inductive step and that of the Lemma
\ref{lem:poly}.

\section{Transformation: MAP in factor graph to binary pair-wise MRF}\label{ap:binary}

In this section we show that any MAP estimation problem is equivalent
to estimating MAP in a specific binary pair-wise problem on a suitably
constructed graph with node potentials. This construction is from
work by Sanghavi, Shah and Willsky\cite{SSW07}.
This construction is related to
the ``overcomplete basis'' representation \cite{WJ03}.  Consider the
following canonical MAP estimation problem: suppose we are given
a distribution $q(\by)$ over vectors $\by = (y_1,\ldots,y_M)$ of
variables $y_m$, each of which can take a finite value. Suppose
also that $q$ factors into a product of strictly positive functions,
which we find convenient to denote in exponential form:
\[
q(\by) ~ = ~ \frac{1}{Z} \prod_{\alpha \in A} \exp \lf (
\phi_\alpha(\by_\alpha) \rf ) ~ = ~ \frac{1}{Z} \exp \lf (
\sum_{\alpha\in A} \phi_\alpha(\by_\alpha) \rf )
\]
Here $\alpha$ specifies the domain of the function $\phi_\alpha$, and
$\by_\alpha$ is the vector of those variables that are in the domain
of $\phi_\alpha$. The $\alpha$'s also serve as an index for the
functions. $A$ is the set of functions. The MAP estimation problem is
to find a maximizing assignment $\by^* \in \arg \max_\by q(\by)$.

We now build an auxillary graph $\tG$, and assign weights to its
nodes, such that the MAP estimation problem above is equivalent to
finding the MWIS of $\tG$.  There is one node in $\tG$ for each pair
$(\alpha,\by_\alpha)$, where $\by_\alpha$ is an {\em assignment}
(i.e. a set of values for the variables) of domain $\alpha$. We will
denote this node of $\tG$ by $\delta(\alpha,\by_\alpha)$.

There is an edge in $\tG$ between any two nodes
$\delta(\alpha_1,\by_{\alpha_1}^1)$ and
$\delta(\alpha_2,\by_{\alpha_2}^2)$ if and only if there exists a
variable index $m$ such that
\begin{enumerate}
\item $m$ is in both domains, i.e. $m\in \alpha_1$ and $m\in
  \alpha_2$, and
\item the corresponding variable assignments are different,
  i.e. $y^1_m \neq y^2_m$.
\end{enumerate}
In other words, we put an edge between all pairs of nodes that
correspond to {\em inconsistent} assignments.  Given this graph $\tG$,
we now assign weights to the nodes. Let $c>0$ be any number such that
$c+\phi_\alpha(\by_\alpha) > 0$ for all $\alpha$ and $\by_\alpha$. The
existence of such a $c$ follows from the fact that the set of
assignments and domains is finite. Assign to each node
$\delta(\alpha,\by_\alpha)$ a weight of $c+\phi_\alpha(\by_\alpha)$.
Consider an example of this construction first. Later, we state the
precise equivalence.

\begin{figure}[th]
\begin{center}
\begin{psfrags}
\includegraphics[width=5cm]{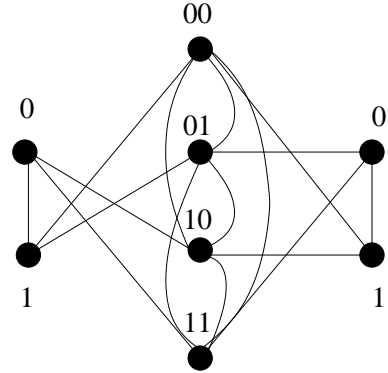}
\end{psfrags}
\end{center}
\caption{Example of transforming MAP for factor graph to MAP in binary
pair-wise MRF.}
\label{fig:map2mwis} \vspace{-.1in}
\end{figure}

\begin{example}
Let $y_1$ and $y_2$ be binary variables with joint distribution
\[
q(y_1,y_2) ~ = ~ \frac{1}{Z}\exp(\theta_1 y_1 + \theta_2 y_2 +
\theta_{12}y_1 y_2)
\]
where the $\theta$ are any real numbers. The corresponding $\tG$ is
shown in Figure \ref{fig:map2mwis}. Let $c$ be any number such that $c+\theta_1$,
$c+\theta_2$ and $c+\theta_{12}$ are all greater than 0. The weights
on the nodes in $\tG$ are: $\theta_1 + c$ on node ``1'' on the left,
$\theta_2 + c$ for node ``1'' on the right, $\theta_{12}+c$ for the
node ``11'', and $c$ for all the other nodes.
\end{example}

\begin{lemma}\label{lem:map2mwis}
Suppose $q$ and $\tG$ are as above.
(a) If $\by^*$ is a MAP estimate of $q$, let $\delta^* = \{
  \delta(\alpha,\by^*_\alpha)\,|\, \alpha\in A\}$ be the set of nodes
  in $\tG$ that correspond to each domain being consistent with
  $\by^*$. Then, $\delta^*$ is an MWIS of $\tG$.
(b) Conversely, suppose $\delta^*$ is an MWIS of $\tG$. Then, for
  every domain $\alpha$, there is exactly one node
  $\delta(\alpha,\by^*_\alpha)$ included in $\delta^*$.  Further, the
  corresponding domain assignments$\{\by^*_\alpha\,|\,\alpha\in A\}$
  are consistent, and the resulting overall vector $\by^*$ is a MAP
  estimate of $q$.
\end{lemma}
\begin{proof}
A {\em maximal} independent set is one in which every node is either
in the set, or is adjacent to another node that is in the set. Since
weights are positive, any MWIS has to be maximal. For $\tG$ and $q$
as constructed, it is clear that
\begin{enumerate}
\item If $\by$ is an assignment of variables, consider the
  corresponding set of nodes $\{\delta(\alpha,\by_\alpha)\,|\,
  \alpha\in A\}$. Each domain $\alpha$ has exactly one node in this
  set. Also, this set is an independent set in $\tG$, because the
  partial assignments $\by_\alpha$ for all the nodes are consistent
  with $\by$, and hence with each other. This means that there will
  not be an edge in $\tG$ between any two nodes in the set.
\item Conversely, if $\Delta$ is a maximal independent set in $\tG$,
  then all the sets of partial assignments corresponding to each node
  in $\Delta$ are all consistent with each other, and with a global
  assignment $\by$.
\end{enumerate}
There is thus a one-to-one correspondence between maximal independent
sets in $\tG$ and assignments $\by$. The lemma follows from this
observation.
\end{proof}





\end{document}